\documentclass[preprint,12pt]{elsarticle}

\journal{Robotics and Autonomous Systems}

\usepackage[table,xcdraw,dvipsnames]{xcolor}

\usepackage{amsmath}
\usepackage{amssymb}
\usepackage{amsfonts}
\usepackage{orcidlink}
\usepackage{booktabs}

\usepackage{colortbl}
\usepackage{array}
\usepackage{tikz}
\usepackage{siunitx}
\usepackage{multirow}
\usepackage{bm}
\usepackage{bbm}
\usepackage{cases}
\usepackage[ruled,vlined]{algorithm2e}
\usepackage{algorithmicx}
\usepackage{algpseudocode}
\usepackage[maxfloats=256]{morefloats}
\usepackage{mathtools}
\usepackage[whole]{bxcjkjatype}
\usepackage{url}
\usepackage{float}

\usepackage{hyperref}
\usepackage{caption}
\usepackage{mathtools}

\definecolor{customgreen}{RGB}{0,200,0}
\newcommand{\greencheck}{\textcolor{customgreen}
{\checkmark}}

\SetKwInput{KwInput}{Input}
\SetKwInput{KwOutput}{Output}
\SetKwInput{KwInit}{Init}

\usetikzlibrary{positioning}
\usetikzlibrary{arrows,shapes,automata,backgrounds,fit,petri,calc}

\tikzstyle{latent} = [circle,fill=white,draw=black,inner sep=1pt,
minimum size=25pt, font=\fontsize{10}{10}\selectfont, node distance=1]

\tikzstyle{pseud} = [minimum size=25pt, font=\fontsize{10}{10}\selectfont, node distance=1]

\colorlet{DPNColor}{orange}
\colorlet{RCANColor}{RubineRed}

\newcommand{\mA}{\mathbf{A}} 
\newcommand{\mI}{\mathbf{I}} 
\newcommand{\mR}{\mathbf{R}} 
\newcommand{\vt}{\mathbf{t}} 

\newcommand{\mRrod}{\mR_{\textrm{rod}}} 

\newcommand{\vp}{\mathbf{p}} 
\newcommand{\vq}{\mathbf{q}} 
\newcommand{\vn}{\mathbf{n}} 
\newcommand{\vc}{\mathbf{c}} 
\newcommand{\vb}{\mathbf{b}} 
\newcommand{\vv}{\mathbf{v}} 
\newcommand{\vx}{\mathbf{x}} 
\newcommand{\vy}{\mathbf{y}} 
\newcommand{\va}{\mathbf{a}} 
\newcommand{\vu}{\mathbf{u}} 

\newcommand{\vpp}{\mathbf{p}'} 
\newcommand{\vzero}{\mathbf{0}} 

\newcommand{\dd}{\delta d} 

\newcommand{\ddSigma}{\dd_{\Sigma}}

\newcommand{\mRSigma}{\mR_{\Sigma}}
\newcommand{\vtSigma}{\vt_{\Sigma}}

\newcommand{\vomega}{\bm{\omega}}

\newcommand{\Sf}{\mathcal{S}^f}
\newcommand{\So}{\mathcal{S}^o}

\newcommand{\pF}{\partial \mathcal{F}}
\newcommand{\pO}{\partial \mathcal{O}}

\newcommand{\T}{\mathcal{T}}
\newcommand{\HpO}{\mathcal{H}_{\pO}}
\newcommand{\W}{\mathcal{W}}

\usepackage{makecell}

\newcommand{\subhead}[1]{%
  \par\addvspace{0.6\baselineskip}
  \noindent\textbf{#1.}
}

\newif\ifhighlight
\highlighttrue        

\begin{document}

\begin{frontmatter}

\title{
    DISF: Disentangled Iterative Surface Fitting for Contact-stable Grasp Planning with Grasp Pose Alignment to the Object Center of Mass
}

\author[naist]{Tomoya Yamanokuchi\corref{cor1}\orcidlink{0000-0003-2387-2197}}
\ead{yamanokuchi.tomoya@naist.ac.jp}

\author[padua]{Alberto Bacchin\orcidlink{0000-0002-2945-8758}}
\ead{bacchinalb@dei.unipd.it}

\author[padua]{Emilio Olivastri\orcidlink{0000-0002-0003-5559}}
\ead{emilio.olivastri@phd.unipd.it}

\author[naist]{\\Ryotaro Arifuku\orcidlink{0009-0009-6837-3464}}
\ead{arifuku.ryotaro.au2@naist.ac.jp}

\author[naist]{Takamitsu Matsubara\orcidlink{0000-0003-3545-4814}}
\ead{takam-m@is.naist.jp}

\author[padua]{Emanuele Menegatti\orcidlink{0000-0001-5794-9979}}
\ead{emg@dei.unipd.it}

\cortext[cor1]{Corresponding author.}

\affiliation[naist]{%
    organization={
        Division of Information Science, Graduate School of Science and Technology, Nara Institute of Science and Technology
    }, 
    addressline={8916-5 Takayama}, 
    city={Ikoma}, 
    state={Nara}, 
    postcode={630-0192}, 
    country={Japan}
}

\affiliation[padua]{%
    organization={
        Department of Information Engineering, University of Padua
    }, 
    addressline={\\Via Gradenigo 6/b}, 
    city={Padua}, 
    postcode={35131}, 
    country={Italy}
}

\begin{abstract}
    In this work, we address the limitation of surface fitting-based grasp planning algorithm, which primarily focuses on geometric alignment between the gripper and object surface while overlooking the stability of contact point distribution, 
    often resulting in unstable grasps due to inadequate contact configurations. 
    To overcome this limitation, we propose a novel surface fitting algorithm that integrates contact stability while preserving geometric compatibility.  
    Inspired by human grasping behavior, our method disentangles the grasp pose optimization into three sequential steps: (1) rotation optimization to align contact normals, (2) translation refinement to improve the alignment between the gripper frame origin and the object Center of Mass (CoM), and (3) gripper aperture adjustment to optimize contact point distribution.
    We validate our approach in simulation across 15 objects under both Known-shape (with clean CAD-derived dataset) and Observed-shape (with YCB object dataset) settings, including cross-platform grasp execution on three robot--gripper platforms.
    We further validate the method in real-world grasp experiments on a UR3e robot.
    Overall, DISF reduces CoM misalignment while maintaining geometric compatibility, translating into higher grasp success in both simulation and real-world execution compared to baselines.
    Additional videos and supplementary results are available on our project page: https://tomoya-yamanokuchi.github.io/disf-ras-project-page/
\end{abstract}


\end{frontmatter}

\section{Introduction}    
    
    Observations of human grasping behavior suggest that aligning the Center of Mass (CoM) of the hand closer with that of the object improves grasp stability~\cite{CoM1,CoM2,CoM3}.
    This is because CoM misalignment induces large rotational moment, which can destabilize the grasp~\cite{moment}.   
    Following this principle, numerous bio-inspired algorithms have been proposed to determine optimal grasp configuration~\cite{biomechanics_ellipsoid_1992,biomechanics_JoB_2005}. 
    However, while these algorithms achieve high accuracy in predicting human grasping behavior, they are fundamentally limited by the assumption that objects can be represented using simple geometric models, such as cylinders~\cite{biomechanics_EMBC_2011,biomechanics_IJIDeM_2012} or spheres~\cite{biomechanics_BioRob_2014}, making them unable to generalize to complex geometries.

    To overcome these limitations, grasp planning algorithms that do not rely on mathematical models of object shapes have been proposed in recent years, utilizing point cloud data~\cite{Fan_Case2018}.
    This approach builds upon the framework of 3D point cloud registration, which has been well established in the field of Computer Vision~\cite{ICP_TPAM_1987,ICP_point2plane_2001,makadia2006fully,yang2015go,yang2020teaser,vizzo2023kiss}. 
    By representing both the object and the robot hand's gripper surface as point cloud data, and directly optimizing their \textit{geometric compatibility} as an evaluation metric, this method determines an appropriate grasp pose.
    A series of studies by Fan et al. have demonstrated the effectiveness of this geometric compatibility-based optimization approach for grasp planning across a wide range of object shapes~\cite{Fan_IROS_2018,Fan_IROS_2019,Fan_RAL_2019,Fan_Sensors_2024}.

    While surface fitting algorithms based on geometric compatibility optimization offer high flexibility, they do not sufficiently account for whether the aligned surfaces actually lead to a stable grasp. 
    Specifically, achieving a stable grasp requires the ability to generate contact forces that can fully counteract external forces and torques (known as force-closure property~\cite{force_closure}). 
    However, by focusing solely on geometric alignment, these methods fail to consider fundamental factors necessary for generating contact forces, such as the appropriate spatial relationship between the hand and the object. 
    As a result, even if the surfaces are geometrically well-aligned, a spatial gap can form between the hand and the object, preventing actual contact from being established, or leading to an unstable distribution of contact points. 
    
\begin{figure}[t]
        \centering
        \includegraphics[width=\columnwidth]{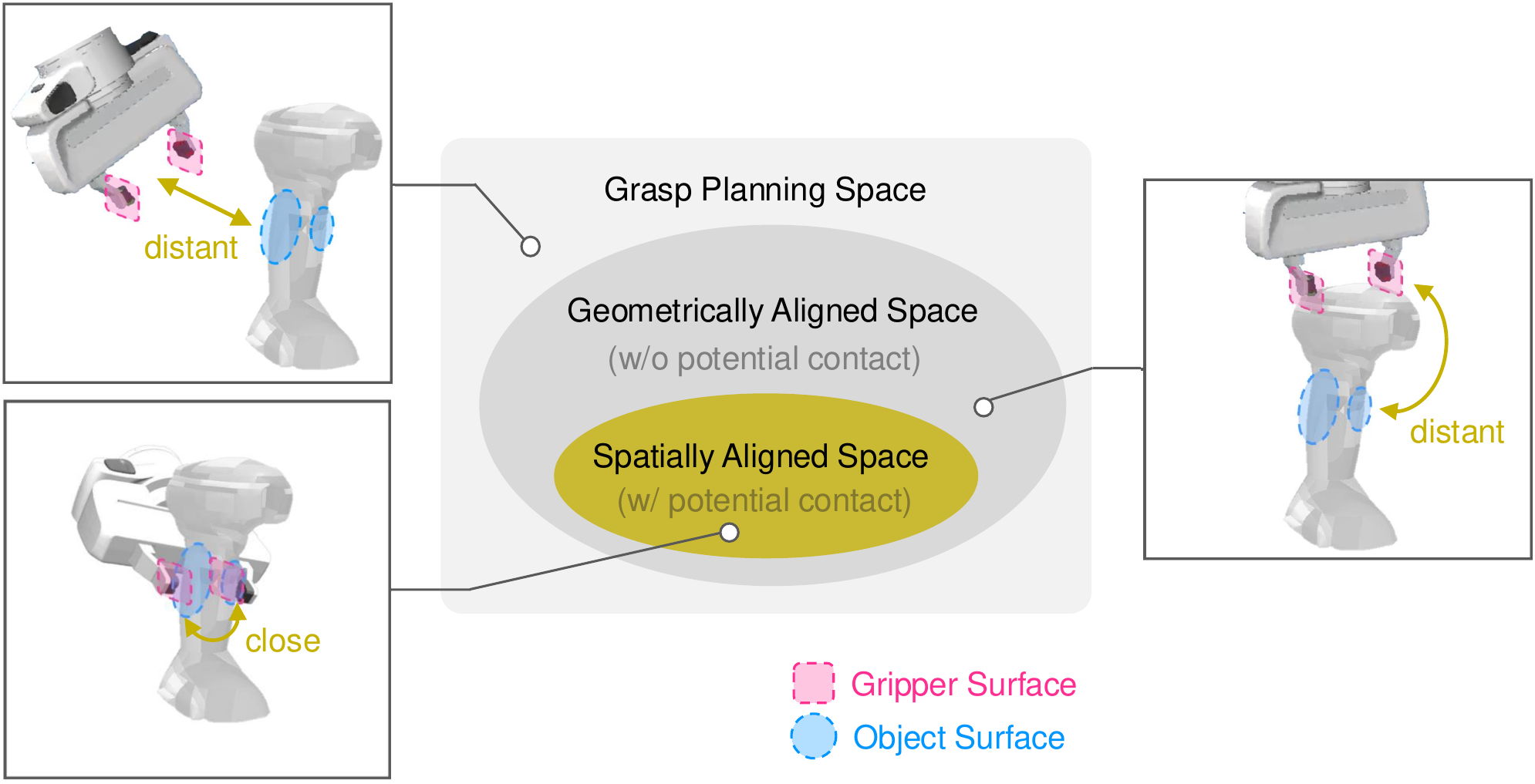}
        \caption{   
            The relationship between the grasp planning space, geometrically aligned space, and spatially aligned space.  
        }        \label{fig:relationship_of_contact_stablegrasp_space}
    \end{figure}

    To address this issue, it is essential not only to align surfaces based on geometric compatibility but also to ensure that the robotic hand (or gripper) and object surfaces are spatially well-aligned, allowing contact to be potentially established. 
    We refer to this spatial alignment, which facilitates contact, as \textit{contact stability} (Fig.~\ref{fig:relationship_of_contact_stablegrasp_space}).

    In this study, we propose a novel surface fitting-based grasp planning algorithm that incorporates contact stability alongside geometric compatibility, which we call Disentangled Iterative Surface Fitting (DISF). 
    From the perspective of contact stability, we explicitly integrate CoM alignment into the optimization process, drawing inspiration from the observation that, as mentioned earlier, humans naturally align their hand's CoM with that of the object to enhance grasp stability~\cite{CoM1,CoM2,CoM3}. 
    To achieve this, we leverage another key insight from human grasping behavior-that different pose parameters are optimized sequentially rather than simultaneously~\cite{human_grasp_analysis_2014}-and disentangle the overall grasp pose optimization into the following three sequential stages:
    (1) rotation optimization to align contact normals,  
    (2) translation refinement for CoM alignment, and  
    (3) gripper aperture adjustment to optimize contact point distribution.   
    Our disentangled optimization framework preserves the advantages of flexible geometric compatibility evaluation while systematically enhancing contact stability through CoM alignment. The overview of our DISF framework is shown in Fig.~\ref{fig:disf_algprithm}.

    \begin{figure}[t]
        \centering
        \includegraphics[width=\columnwidth]{figures/disf_concept_v6-crop.pdf}
        \caption{   
            Overview of the proposed DISF optimization process. The grasp pose optimization is disentangled into three sequential steps: (1) rotation optimization to align contact normals, (2) translation refinement for Center of Mass (CoM) alignment, and (3) gripper aperture adjustment to optimize contact point distribution. Each step iteratively updates the gripper transformation parameters to ensure both geometric compatibility and contact stability. The arrows indicate the optimization flow, illustrating how the gripper adapts to the object surface through iterative surface fitting.
        }
        \label{fig:disf_algprithm}
    \end{figure}

    To evaluate the effectiveness of DISF, we conducted a comprehensive study in both simulation and the real world.
    In simulation, we first quantified grasp quality on 15 objects spanning Known-shape and Observed-shape settings using two complementary criteria: geometric compatibility (geometric misalignment) and contact stability (CoM misalignment).
    We then assessed physical feasibility by executing the planned grasps in a physics simulator across three robot--gripper platforms, and reported grasp success rates for each setting.
    Finally, we evaluated real-world grasp execution using observed object geometry reconstructed from two depth sensors, demonstrating the practical benefit of CoM alignment in surface-fitting-based grasp planning.

    This paper substantially extends our preliminary conference paper~\cite{IAS-DISF} by (i) expanding the simulation study to cover both Known-shape (clean CAD-derived point clouds) and Observed-shape (YCB-derived point clouds with sensor/reconstruction artifacts) settings, (ii) evaluating grasp executions across multiple robot--gripper platforms to assess cross-platform generalization, and (iii) adding real-world grasp experiments on a UR3e robotic platform using observed point cloud data.

    Our main contributions were summarized as follows:
    \begin{itemize}
        \item We proposed DISF, a surface fitting-based grasp planner that integrated contact stability with geometric compatibility via a disentangled, sequential optimization over rotation, translation (CoM alignment), and gripper aperture.
        \item We formalized contact stability in the surface fitting context and introduced CoM misalignment as a complementary grasp quality measure alongside geometric misalignment.
        \item We evaluated the proposed method through (i) grasp quality analysis on 15 objects under both Known-shape and Observed-shape settings, (ii) cross-platform grasp success rates on three robot--gripper platforms in simulation, and (iii) real-world grasp success rates on a UR3e robot with observed object shapes reconstructed from two depth sensors.
    \end{itemize}

\section{Related Works}
    \subsection{
        Center of Mass influence on Human Grasping Behavior
    }

    A stable human grasp is often achieved by bringing the hand's CoM close to the object's CoM, as this tends to increase the contact area, regularize the distribution of forces at contact, and mitigate rotational moments. Previous work suggests that humans can predict the CoM of an object and adapt the hand's grasping pose accordingly. For example, Lukos et al. examined the selection of contact-points in 12 participants under conditions where the CoM of the object was provided or unknown, and found that participants systematically shifted their grasp contacts when the CoM could be predicted~\cite{CoM1}. Desanghere and Marotta further examined CoM estimation from visual cues and reported that gaze fixation is sensitive to CoM location, influencing subsequent grasp placement~\cite{CoM2}. These results underscore the central importance of CoM alignment in human grasping. Building on this insight, we investigate how to incorporate CoM alignment in a principled way into computational grasp planning, with a particular focus on our surface-fitting-based method.
    
    After highlighting the importance of CoM alignment in human grasp selection and stability, an open question remains: how should this principle be encoded in computational grasp models? Biomechanical evidence suggests that CoM alignment can emerge implicitly from the optimization of joint configurations and contact locations. In the model proposed by Lee et al.~\cite{biomechanics_JoB_2005}, stable contacts arise as the finger joints conform to the object surface, and CoM alignment is achieved as a by-product of this optimization, reducing net moments acting on the object. In parallel, computational approaches have introduced CoM-aware grasp prediction models that enforce CoM alignment more explicitly. For example, Klein et al. showed that shifting an object's CoM leads to systematic changes in predicted grasp locations, supporting the hypothesis that CoM-sensitive optimization can explain key aspects of human grasping behavior~\cite{CoM3}. Building on these insights, we integrate CoM alignment into a surface-fitting-based grasp planning framework for robotic grasping, enabling the planner to account not only for geometric compatibility with the object surface, but also for contact stability.

    \subsection{
        Iterative Surface Fitting in Grasp Planning
    }
        Surface fitting methods in robotic grasp planning are based on the idea that a good grasp pose brings the hand's fingertip surfaces into a geometric alignment with the object surface~\cite{biomechanics_JoB_2005,taxonomy_TRO_1989,taxonomy_THMS_2016}. This alignment tends to increase the effective contact area, improving force transmission and reducing the likelihood of slip. As a result, surface alignment provides a practical criterion for generating stable grasps in robotic manipulation. Building on these assumptions, grasp planning can be cast as a point-cloud alignment problem, as shown in recent work~\cite{Fan_Case2018}. In this setting, the objective is to optimize the alignment between the gripper surface and a point cloud representing the object, avoiding the need for an explicit mathematical model of the object geometry, a relevant limitation of many traditional biomechanics-based grasp prediction approaches. Consequently, surface fitting provides a flexible grasp planning framework that is not restricted to predefined geometric primitives and can, in principle, be applied to objects of arbitrary shape~\cite{Fan_IROS_2018,Fan_IROS_2019,Fan_RAL_2019}.
    
        However, most existing surface-fitting-based methods primarily address geometric compatibility and often omit additional stability-related constraints, such as the configuration of contact points with respect to the object CoM. As a result, the planned grasps may be geometrically valid yet mechanically unstable. In this work, we incorporate CoM alignment into surface-fitting-based grasp planning, enabling the optimization to account for both surface alignment and contact stability, and thereby promoting more robust grasp configurations.


\section{Preliminaries}
\label{sec:preliminars}
    \subsection{Contact Surface Optimization}
\begin{figure}[thpb]
        \centering
        \includegraphics[width=\columnwidth]{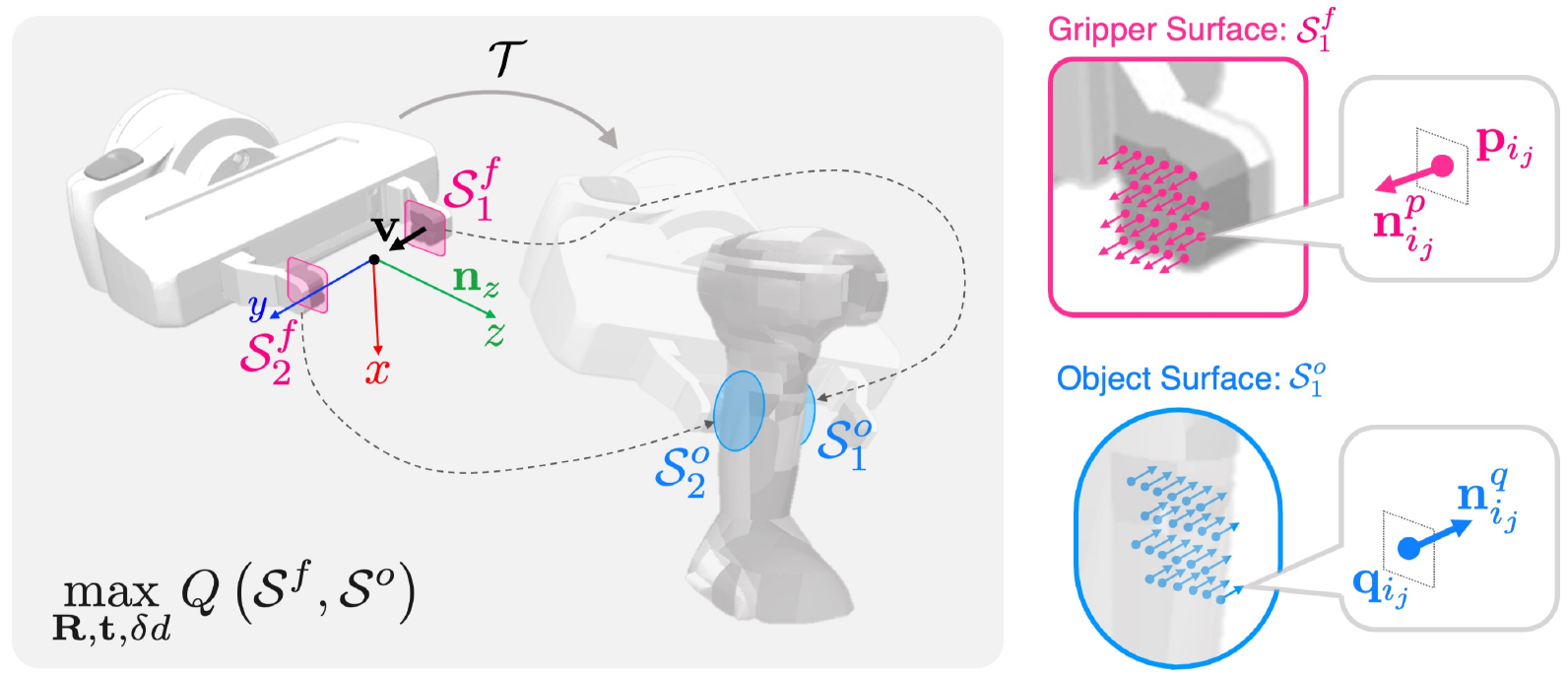}
        \caption{
            Contact surface optimization formulation for antipodal grasp planning.
            Given the canonical fingertip surfaces, we define finger contact surface patches $\Sf_1$ and $\Sf_2$ and transform them by $\T(\cdot;\mR,\vt,\delta d)$, where $(\mR,\vt)$ and the fingertip displacement $\delta d$ are the optimization variables.
            For each transformed patch, the corresponding object contact surface $\So_j$ is obtained by the correspondence function $\HpO(\Sf_j)$ on the object surface.
            Surface samples and normals $\{\vp_{i_j}, \vn^p_{i_j}\}$ and $\{\vq_{i_j}, \vn^q_{i_j}\}$ are used to evaluate the grasp quality $Q(\Sf,\So)$, which is maximized under the gripper workspace and aperture constraints.
        }
        \label{fig:contact_surface_opt}
    \end{figure}

        The grasp planning problem with antipodal grippers can be modeled as a contact surface optimization problem which maximizes the grasp quality $Q$ by optimizing the rotation and translation parameter ($\mR, \vt$) as well as the fingertip displacement $\dd$ from the original gripper aperture $d$, given a specific set of contact surfaces between the fingertip and object:
        \begin{subequations}
        	\label{eq:general_form}
        	\begin{align}
        	\max_{\mR, \vt, \dd} &\  Q(\Sf, \So) \label{eq1:cost}\\
        	\mathrm{s.t.} \quad 
                & \Sf_j \subset \T(\pF_j, \mR, \vt, \dd), \quad j = 1,2 \label{eq1:surface_finger}\\
        	& \So_j = \HpO(\Sf_j), \quad j = 1,2 \label{eq1:surface_object}\\
        	& \Sf_j \in \W(d_0 + \delta d) \quad j = 1,2 \label{eq1:constraint1}\\
        	& d_0 + \dd \in [d_{\text{min}}, d_{\text{max}}] \label{eq1:constraint2}
        	\end{align}
        \end{subequations}        
        where $j \in \{1, 2\}$ is the finger index, $\Sf_j$ and $\So_j$ are the finger and object contact surfaces. 
        The $\Sf = [\Sf_1, \Sf_2]$ is the set of contact surfaces across the multiple fingers and the $\So = [\So_1, \So_2]$ is the corresponding set of contact surfaces for the object. 
        The finger contact surface lies on its canonical surface $\pF_j$ projected by the transformation function $\T$. 
        The object contact surface $\So_j$ is determined by a correspondence matching algorithm $\HpO$ given the object canonical surface $\pO$ and its query $\Sf_j$. 
        The contact surfaces are constrained by the gripper's working range \([d_{\text{min}}, d_{\text{max}}]\), defined by the robot's kinematics. The optimization aims to determine the optimal gripper transformation \((\mathbf{R}^*, \mathbf{t}^*, \delta d^*)\).
        This concept of contact surface optimization is illustrated in Fig.~\ref{fig:contact_surface_opt}. 

    \subsection{Iterative Surface Fitting}
        This section instantiates the contact surface optimization in Eq.~\eqref{eq:general_form} with an iterative surface fitting procedure.
        As illustrated in Fig.~\ref{fig:contact_surface_opt}, we optimize the gripper transformation parameters by iteratively aligning fingertip surface samples to their corresponding object samples.
        
        \subsubsection{Gripper Transformation}
            Given a specific contact point-normal pair $(\vp_{i_j}, \vn_{i_j}^p) \in \Sf_j$, where $i = 1, \dots, N$ and $N$ is the number of the pair for fingertip contact surfaces, the transformation function $\T$ for the gripper is defined as follows:
            \begin{equation}
                \label{eq:gripper_point_normal_set_transformation}
                \begin{aligned}
                    \T ((\vp_{i_j}, \vn_{i_j}^p), \mR(\vomega), \vt, \dd)  = (\mR(\vomega) \vp_{i_j} + \vt + 0.5(-1)^j {\mR(\vomega)}{\vv \dd}, \mR(\vomega) \vn_{i_j}^p)
                \end{aligned}
            \end{equation}
            where $\mR(\vomega)$ is the rotation matrix parameterized with the axis-angle vector $\vomega \in so(3)$, $\vv \in \mathbb{R}^3 $ is the unit vector pointing from $\Sf_1$ to $\Sf_2$. 

        \subsubsection{Grasp Quality Measures}
            \label{subsubsec:grasp_quality_measures}
            The geometric compatibility of point cloud data is generally evaluated using surface distance, called point-to-plane distance~\cite{ICP_point2plane_2001}.
            Therefore, grasp quality in surface fitting-based grasp planning is also assessed based on this criterion~\cite{Fan_Case2018,Fan_IROS_2018,Fan_IROS_2019}, which is defined as the distance between each point on the gripper surface and the tangent plane of the object surface:
            \begin{equation}
                \label{eq:Ep}
                \begin{aligned}
                     E_p(\vomega,\vt,\dd; \Sf, \So) &  = \sum_{i=1}^{N} \sum_{j = 1}^{2}\left((\vpp_{i_j}-\vq_{i_j})^\top \vn_{i_j}^q \right)^2, 
                \end{aligned}
            \end{equation}
            where $\vq_{i_j}$ and $\vn^q_{i_j}$ are the point and normal vector on the object contact surface $\So$. 
            
            In the context of grasp planning, normal misalignment is also an important criterion, as it directly affects the ability to achieve force-closure properties~\cite{force_closure}. 
            Stable contact requires the normals of the gripper and object surfaces to be oriented in opposite directions; thus, misalignment is measured by evaluating the deviation between these two normal vectors.
            \begin{equation}
                \label{eq:En}
                \begin{aligned}
                     E_n(\vomega; \Sf, \So)   = \sum_{i = 1}^N \left(\left(\mR(\vomega) \vn_{i}^p \right)^\top \vn_{i}^q + 1 \right)^2.
                \end{aligned}
            \end{equation}
            
            Additionally, approach direction misalignment~\cite{Fan_Sensors_2024}, is also often used. 
            In this study, we adopt this concept with modifications to better suit our formulation: 
            \begin{equation}
                \label{eq:Ea}
                \begin{aligned}
                     E_a(\vomega; \Sf, \So)   = \sum_{i = 1}^N \left(\left(\mR(\vomega) \vn_z \right)^\top \vn_{app} - 1\right)^2,
                \end{aligned}
            \end{equation}
            where $\vn_z$ is the z-axis direction of the gripper defined by the hand plane, and $\vn_{app}$ is the approach direction of the gripper. 
            This metric plays a crucial role in avoiding collisions and achieving a natural grasping posture.

        \subsubsection{Gradient-based Optimization with Iterative Least-squares Method}
            \label{sec:method:optimization}
            Direct optimization of the rotation matrix is challenging due to its constraint on the special orthogonal group $SO(3)$. 
            To address this, we approximate it as a small rotation around each axis~\cite{small_rotation}, allowing the rotation matrix to be expressed as:
            \begin{equation}
                \label{eq:R_skew}
                \mR(\vomega) \approx \mI + [\vomega]_{\times}
            \end{equation}
            where $\mI$ is $3 \times 3$ identity matrix, $[\cdot]_{\times}$ means skew-symmetric matrix, and $[\vomega]_{\times}$ is skew-symmetric matrix with the small rotation vector $\vomega = [\omega_x, \omega_y, \omega_z]^\top$: 
            \begin{equation}
                \label{eq3:closed_form}
                [\vomega]_{\times} = \left[\begin{array}{ccc}
                0         & \omega_z & \omega_y \\
                \omega_z  &  0       & -\omega_x \\
                -\omega_y & \omega_x & 0
                \end{array}\right].  
            \end{equation}
             This linearization simplifies the optimization process by reducing the complexity of handling the full rotation matrix $\mR$.  By applying the small rotation approximation to the grasp quality measures, the rotational component can be linearized into a form that is compatible with least-squares optimization, as follows:
            \begin{equation} 
            \label{eq:least_squares} 
                \min_{\vx} \|\mA \vx - \vb\|^2. 
            \end{equation}
            where $\vx$ contains the unknown parameters, e.g. $\vomega$, $\vt$, or $\dd$, $\mA$ is the coefficient matrix derived from the Jacobians of the grasp quality measures with respect to the parameters in $\vx$, and $\vb$ is the residual vector. 
            Taking the derivative of the squared error with respect to \( \vx \) and setting it to zero yields the normal equation $\mA^\top \mA \vx = \mA^\top \vb$ whose solution is given by $\vx = (\mA^\top \mA)^{-1} \mA^\top \vb$. 
            This equation is solved iteratively to maximize the grasp quality measures.

\section{Proposed Methods}
    \label{sec:method}
    
    Starting from the surface fitting formulation in Section~\ref{sec:preliminars}, our goal is to improve the optimization process introducing contact-stability awareness. Incorporating such a term directly into the joint surface-fitting objective, however, is non-trivial. Motivated by evidence that key posture parameters (i.e., rotation, translation, and aperture) are optimized sequentially rather than jointly in human grasping~\cite{human_grasp_analysis_2014}, we propose to disentangle grasp-pose optimization into three stages instead of solving for all parameters simultaneously: 
    (1) rotation optimization to align contact normals, 
    (2) translation refinement to promote CoM alignment, and 
    (3) gripper-aperture adjustment to improve the distribution of contact points. 
    This staged formulation enables us to introduce additional constraints that would be difficult to incorporate in a single entangled optimization problem.

    In the following, we detail each disentangled optimization stage and then present the complete Disentangled Iterative Surface Fitting (DISF) algorithm.

    \subsection{Rotation Optimization for Contact Normal Misalignment (RotOpt)}
            \begin{algorithm}[t]
                \caption{Rotation Optimization for Contact Normal Misalignment (RotOpt)}
                \label{alg:PRO-CAM}
                \begin{algorithmic}[1]
                    \State \textbf{Input:} $\pF, \Sf, \So, \mR^v, \vv$
                    \State $\vomega^* \gets \underset{\vomega}{\mathrm{min}} \, E_{na} \left(\vomega; \Sf, \So \right)$
                    \State $\mR^v \gets \mRrod(\vomega^*) \mR^v$
                    \State $\vv \gets \mR^v \vv$
                    \State $\Sf_j \gets \T(\Sf_j, \mRrod(\vomega^*), \vt=\vzero, \dd=0), \quad j=1,2$
                    \State $\pF_j \gets \T(\pF_j, \mRrod(\vomega^*), \vt=\vzero, \dd=0), \quad j=1,2$
                    \State \Return{$(\pF, \Sf, \vomega^*, \mR^v, \vv)$}
                \end{algorithmic}
            \end{algorithm}

        The first stage optimizes the rotation parameter $\vomega$ to improve contact-normal alignment, thereby promoting force-closure properties~\cite{force_closure}. We treat this step separately from surface-distance minimization, which has been the primary focus of prior surface-fitting approaches~\cite{Fan_Case2018,Fan_IROS_2018,Fan_IROS_2019}. This staged design is also consistent with human behavior, where hand orientation is often adjusted before translating toward the object. Since rotation additionally influences the approach direction, we define a rotation objective that jointly accounts for contact-normal misalignment and approach-direction misalignment, weighted by a factor $\beta \in \mathbb{R}$:
        \begin{equation}
            \label{eq:E_na}
            E_{na}(\vomega; \Sf, \So) = E_{n}(\vomega; \Sf, \So) + \beta^2 E_{a}(\vomega; \Sf, \So).
        \end{equation}
        The resulting procedure for optimizing the rotation parameters is summarized in Algorithm~\ref{alg:PRO-CAM}.

        Palm rotation optimization can be expressed as a least-squares problem analogous to Eq.~\eqref{eq:least_squares}, using an augmented matrix $\Tilde{\mA} = [\mA_n^\top, \mA_a^\top]^\top$, an augmented residual $\Tilde{\vb} = [\vb_n, \vb_a]^\top$, and the unknown $\vx=\vomega$. Here, $\mA_n = [\va_{n,1}^\top, \dots, \va_{n,N}^\top]^\top$ and $\vb_n = [b_{n,1}, \dots, b_{n,N}]^\top$ correspond to the normal-misalignment term $E_n$, while $\mA_a=\va_a$ and $\vb_a=b_a^\top$ are derived from the approach-direction misalignment:
        \begin{subequations}
        \label{eq:E_na_param}
        \begin{align}
            \va_{n,i} &= (\vn^p_i \times \vn^q_i)^\top, \label{eq:En_param:a}\\
            b_{n,i} &= - \left( (\vn^p_i)^\top \vn^q_i  + 1 \right), \label{eq:En_param:b}\\
            \va_{a} &= \beta (\vn_{z} \times \vn_{app})^\top, \label{eq:Ea_param:a}\\
            b_{a} &= - \beta \left( (\vn_{z})^\top \vn_{app} - 1 \right). \label{eq:Ea_param:b}
        \end{align}
        \end{subequations}

        Once the optimum $\vomega^*$ is obtained, it is mapped to the rotation matrix $\mRrod(\vomega^*)$ via Rodrigues' formula:
        \begin{equation}
        \label{eq:R_rod}
        \mRrod(\vomega) = \mI + \sin\theta [\vu]_{\times} + (1 - \cos\theta) [\vu]_{\times}^2,
        \end{equation}
        where $\mI$ denotes the $3\times 3$ identity matrix, $\vu=\vomega/\|\vomega\|$ is the rotation axis, and $\theta=\|\vomega\|$ is the rotation angle. The resulting rotation is then used to update the current fingertip pointing vector $\vv$ (Lines~3--4), the fingertip contact surfaces $\Sf$ (Line~5), and the corresponding canonical surfaces $\pF$ (Line~6). Finally, the updated parameters are forwarded to the subsequent translation-refinement stage (Line~7).

    \subsection{Translation Refinement for CoM Alignment (TransRefine)}
            \begin{algorithm}[t]
                \caption{Translation Refinement for CoM Alignment (TransRefine)}
                \label{alg:PTR-CEN}
                \begin{algorithmic}[1]
                    \State \textbf{Input:} $\pF, \Sf, \So, \vv$
                    \State $\vc^f \gets \texttt{centroid}(\Sf)$
                    \State $\vc^o \gets \texttt{centroid}(\So)$
                    \State $\vt^c = \vc^o - \vc^f$
                    \State $\Sf_j \gets \T(\Sf_j, \mR=\mI, \vt=\vt^c, \dd=0), \quad j=1,2$
                    \State $\pF_j \gets \T(\pF_j, \mR=\mI, \vt=\vt^c, \dd=0), \quad j=1,2$
                    \State \Return{$(\pF, \Sf, \vt^c)$}
                \end{algorithmic}
            \end{algorithm}

        The second stage refines the translation parameter $\vt$ by promoting alignment between the gripper and object CoMs. The overall procedure is summarized in Algorithm~\ref{alg:PTR-CEN}.

        Because the true CoMs of the object cannot be recovered from surface geometry alone, we approximate them using the \emph{centroids} of the corresponding surfaces. Note that this correspond to assume that the mass distribution is constant over the object. We compute the centroid of an input surface $\partial$ via the \texttt{centroid()} operator:
        \begin{equation}
        \label{eq:centroid}
        \texttt{centroid}(\partial) = \frac{1}{K}\sum_{k=1}^{K} \vy_k, \quad \vy_k \in \partial,
        \end{equation}
        where $K$ denotes the number of points representing the surface (Lines~2--3). 
         
        The resulting translation update $\vt^c$ (Line~4) is then applied to the current gripper surfaces $\Sf$ and $\pF$ (Lines~5--6), and the updated parameters are passed to the subsequent fingertip-displacement optimization stage (Line~7).

    \subsection{Fingertip Displacement Optimization for Stable Contact Distribution (FingerOpt)}

        Given an estimated grasp pose $(\mR(\vomega^*), \vt^*)$, the final stage optimizes the fingertip displacement $\dd$. The complete procedure is summarized in Algorithm~\ref{alg:FingerOpt}. The objective of this step is to refine the gripper aperture so as to reduce the gripper-object surface distance (Eq.~\eqref{eq:Ep}) and, in turn, promote a stable and well-distributed set of contact points. To this end, we solve the following one-dimensional constrained least-squares problem over $\delta d$:
        \begin{equation}
            \min_{\delta d} \sum_{i=1}^N \sum_{j = 1}^{2} (b_{i_j} - a_{i_j}\delta d)^2, \quad 
            \text{s.t.} \quad \delta d + d_0 \in [d_\text{min}, d_\text{max}].
        \end{equation}
        The coefficients are given by
        \begin{subequations}
        \label{eq:Ep_param}
        \begin{align}
            a_{i_j} &= 0.5 (-1)^{j-1} \vv^\top \vn^q_{i_j}, \\
            b_{i_j} &= \left(\vp_{i_j} - \vq_{i_j}\right)^\top \vn^q_{i_j}.
        \end{align}
        \end{subequations}

        For a two-finger parallel gripper, the optimal relative fingertip motion admits a closed-form solution:
        \begin{equation}
        \label{eq:dd_solution_combined}
        \delta d^* =
        \begin{cases} 
        d_\text{min} - d, & \text{if } \delta \hat{d} + d < d_\text{min}, \\
        \delta \hat{d}, & \text{if } d_\text{min} \leq \delta \hat{d} + d \leq d_\text{max}, \\
        d_\text{max} - d, & \text{if } \delta \hat{d} + d > d_\text{max}, \\
        \end{cases}
        \quad 
        \delta \hat{d} = \frac{\sum_{i=1}^m \sum_{j = 1}^{2}a_{ij}b_{ij}}{\sum_{i=1}^m \sum_{j = 1}^{2}a_{ij}^2}.
        \end{equation}

        \begin{algorithm}[t]
            \caption{Fingertip Displacement Optimization for Stable Contact Distribution (FingerOpt)}
            \label{alg:FingerOpt}
            \begin{algorithmic}[1]
                \State \textbf{Input:} $\pF, \Sf, \So, \vv, d$
                \State $\dd^* \gets \underset{\dd}{\mathrm{min}} \, E_p \left(\dd, d; \Sf, \So\right)$
                \State $\Sf_j \gets \T(\Sf_j, \mR=\mI, \vt=\vzero, \dd^*), \quad j=1,2$
                \State $\pF_j \gets \T(\pF_j, \mR=\mI, \vt=\vzero, \dd^*), \quad j=1,2$
                \State \Return{$(\pF, \Sf, \dd^*)$}
            \end{algorithmic}
        \end{algorithm}

    \subsection{Disentangled Iterative Surface Fitting}
        This section introduces the unified surface fitting procedure. The proposed Disentangled Iterative Surface Fitting (DISF) alternates the three optimization stages to iteratively produce feasible grasps that satisfy both geometric compatibility and contact stability. The overall workflow is summarized in Algorithm~\ref{alg:disf}.
        
        DISF begins by initializing the rotation matrix, translation vector, and gripper aperture. Then it repeatedly executes the three stages: (1) RotOpt, (2) TransRefine, and (3) FingerOpt. Among these, only RotOpt relies on the small-rotation approximation, which yields a locally convergent update. By contrast, TransRefine and FingerOpt do not require such approximations and admit closed-form solutions. After each iteration, the updated parameters—namely cumulative rotation $\mRSigma$, translation $\vtSigma$, and fingertip displacement $\ddSigma$—are assessed. The iteration continues until the change in the surface update, quantified as $|e - e_p|$, drops below a predefined threshold $\Delta e$, at which point the algorithm terminates.

    \begin{algorithm}[t]
        \caption{
            DISF: Disentangled Iterative Surface Fitting
        }\label{alg:disf}
        \KwInput{
            $\pF$, 
            $\pO$, 
            $\mR_0$, 
            $\vt_0$, 
            $d_0$, 
            $d_{\textrm{min}}$, 
            $d_{\textrm{max}}$,
            $\vv_0$, 
            $\Delta e$
        }
        \KwInit{
            $\pF_0 \gets \T(\pF, \mR_0, \vt_0, d_0)$,
            $\mRSigma \gets \mI$, 
            $\vtSigma \gets \vzero$, 
            $\ddSigma = 0$,
            $d = d_0$, 
            $\vomega^* = \vzero$, 
            $\vt^* = \vzero$, 
            $\dd^* = 0$, 
            $e_p = \infty$, 
            $e \gets E\left(\Sf, \So\right)$, 
            $\vv = \vv_0$
        }
        \While{
            $e_p - e \geq \Delta e$
        }{
            $e_{p} \gets E\left(\Sf, \So\right)$\;
            $(\pF, \Sf, \vomega^*, \mR^v, \vv) \gets \texttt{RotOpt}(\pF, \Sf, \So, \mR^v, \vv)$\;
            $(\pF, \Sf, \vt^*) \gets \texttt{TransRefine}(\pF, \Sf, \So, \vv)$\;
            $(\pF, \Sf, \dd^*) \gets \texttt{FingerOpt}(\pF, \Sf, \So, \vv, d)$\;
            $d \gets d + \delta d^*$\;
            $e \gets E\left(\Sf, \So\right)$\;
            $\mRSigma \gets \mRrod(\vomega^*) \mRSigma$\;
            $\vtSigma \gets \vt^* + \vtSigma$\;
            $\ddSigma \gets \dd^* + \ddSigma$\;
        }
        $\mR^* = \mRSigma \mR_0 $\;
        $\vt^* = \vtSigma + \vt_0$\;
        $\dd^* = \ddSigma $\;
        \Return{$(\mR^*, \vt^*, \dd^*)$}\;
    \end{algorithm}

\section{Simulation Experiments}
    We evaluated the proposed grasp planner in simulation under two shape-availability settings: \emph{Known-shape}, where clean object point clouds were sampled from CAD meshes, and \emph{Observed-shape}, where noisy partial reconstructions were used.
    Across both settings, we assessed (i) grasp quality in terms of geometric compatibility and CoM alignment, (ii) physical feasibility via grasp execution in physics simulator, and (iii) generality across robot--gripper platforms, with a particular emphasis on gripper variability.
    
    \begin{table}[t]
        \centering
        \caption{
            Gripper geometry specifications used in the cross robot--gripper platform evaluation (Fig.~\ref{fig:robot_platforms}). 
            The $W_{\mathrm{finger}}$ and $H_{\mathrm{finger}}$ denote the fingertip width and height, respectively. All values are shown in millimeters and rounded to integers for consistent formatting.
        }
        \label{tab:gripper_specs}
        \small
        \begin{tabular}{lcccc}
            \toprule
            Gripper & $W_{\mathrm{finger}}$ [mm] & $H_{\mathrm{finger}}$ [mm] & $d_{\min}$ [mm] & $d_{\max}$ [mm] \\
            \midrule
            Franka Hand         & 18   & 18  & 11  & 91 \\
            Robotiq HAND-E      & 20   & 21  & 0   & 50 \\
            UMI gripper         & 119  & 26  & 0   & 80 \\
            \bottomrule
        \end{tabular}
    \end{table}

\begin{figure}[thbp]
    \centering
    \includegraphics[width=\columnwidth]{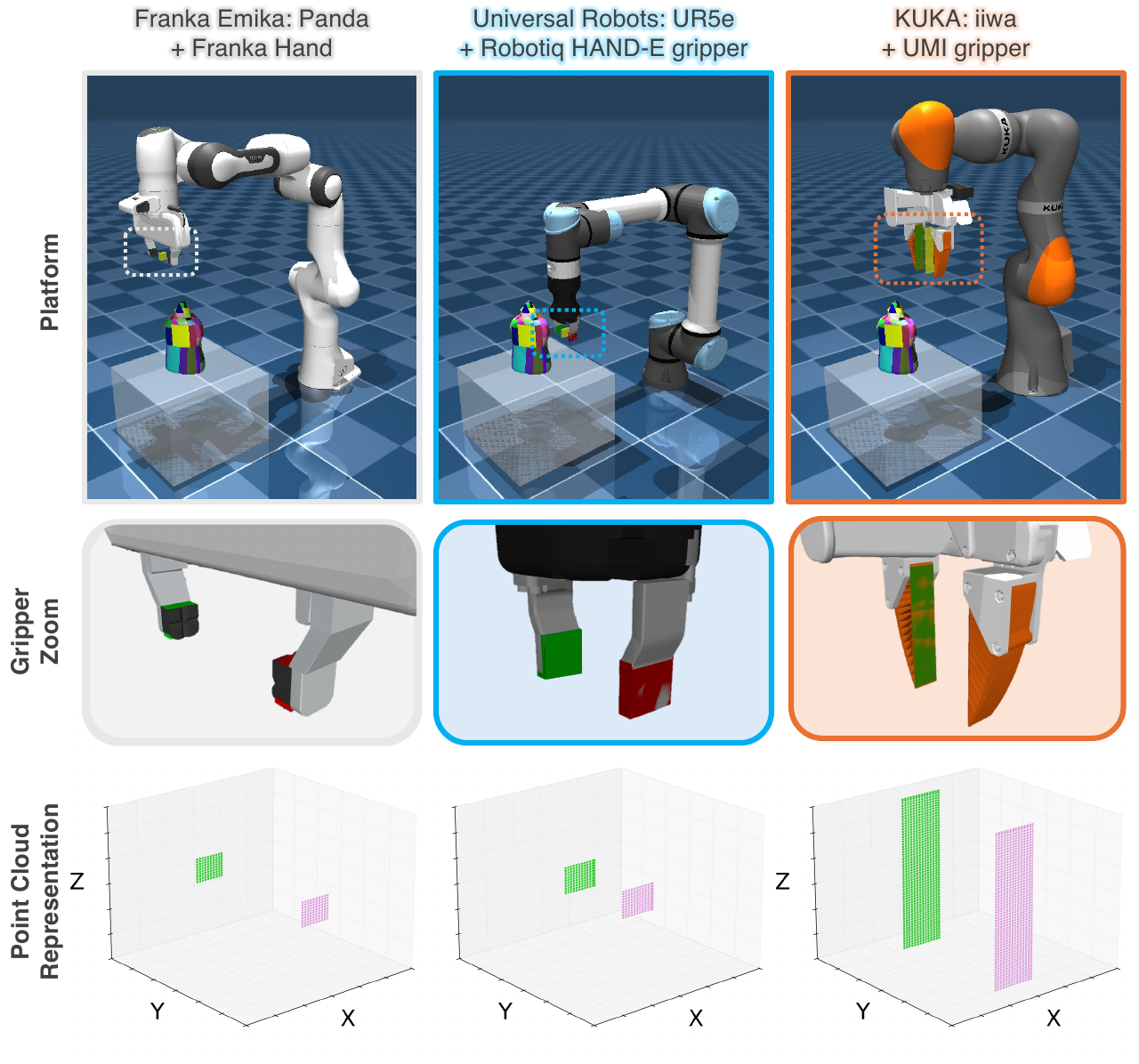}
    \caption{
        Robot--gripper platforms used in simulation for cross-platform evaluation.
        We evaluate grasp execution across three parallel-jaw grippers with different fingertip geometries and aperture ranges:
        \textbf{(left)} Franka Emika Panda with the Franka Hand,
        \textbf{(middle)} Universal Robots UR5e with the Robotiq HAND-E gripper,
        and \textbf{(right)} KUKA iiwa with the UMI gripper~\cite{umi_gripper}.
        The \textbf{Gripper Zoom} row highlights the end-effector designs to emphasize that our evaluation focuses on gripper variability rather than the robot model.
        The \textbf{Point Cloud Representation} row visualizes the corresponding canonical fingertip surfaces used in our surface-fitting optimization.
        In the simulator, the \textbf{left fingertip} is rendered as a \textbf{green} geom and the \textbf{right fingertip} as a \textbf{red} geom; the canonical fingertip point clouds are shown with the same color coding for consistent interpretation.
    }
    \label{fig:robot_platforms}     
\end{figure}

\subsection{Common Setup}
    All simulation experiments were conducted in MuJoCo~\cite{mujoco}.
    Unless otherwise noted, grasp planning and execution were evaluated on the Panda platform equipped with the Franka Hand.
    For cross-platform evaluation, we additionally executed the planned grasps on two other robot--gripper platforms to test robustness to gripper variability, including fingertip-geometry and aperture-range differences. 
    The evaluated robot--gripper platforms are illustrated in Fig.~\ref{fig:robot_platforms}, and the corresponding gripper geometry specifications are summarized in Table~\ref{tab:gripper_specs}.

    \subsubsection{Object Point Clouds}
        We used 15 objects in total: 5 \emph{Known-shape} objects from internet data and 10 \emph{Observed-shape} objects from YCB dataset.
        
\begin{figure}[t]
        \centering
        \includegraphics[width=\columnwidth]{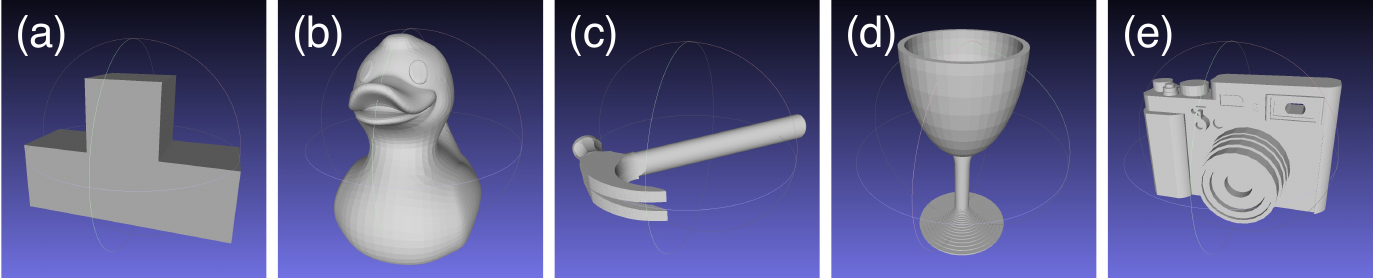}
        \caption{
            Known-shape objects used in simulation, visualized in MeshLab: (a) \texttt{T-shape\_Block}, (b) \texttt{Rubber\_Duck}, (c) \texttt{Hammer}, (d) \texttt{Wine\_Glass}, and (e) \texttt{Old\_Camera}.
        }
        \label{fig:sim_known_shape_objects}
    \end{figure}

        \subhead{Known-shape objects}
            For the Known-shape setting, we downloaded \href{https://www.thingiverse.com/}{CAD meshes} for the five objects and generated clean object point clouds by sampling from the mesh surface using Poisson-disk sampling (Open3D function of \texttt{sample\_points\_poisson\_disk}) with 3000 points.
            The five CAD models used in this setting are shown in Fig.~\ref{fig:sim_known_shape_objects}.
        
        \subhead{Observed-shape objects}
            For the Observed-shape setting, we selected 10 objects from the YCB dataset~\cite{YCB_Dataset} and constructed object point clouds by voxel downsampling with voxel size 5 [mm], which yielded noisier and less complete geometry compared to the Known-shape setting.
            To avoid confusion, we emphasize that our \textit{Observed-shape} setting does \emph{not} assume access to a clean CAD model.
            Instead, we use object geometry that is \emph{reconstructed from real RGB-D observations} provided by the Yale--CMU--Berkeley (YCB) object and model set~\cite{YCB_Dataset}, which includes RGB-D scans, point-cloud data, and reconstructed meshes of physical objects.
            As a result, the observed point clouds can exhibit measurement noise, occlusions, missing depth, and surface reconstruction artifacts (e.g., depth failure on transparent/reflective regions), as described in \cite{YCB_Dataset}, making them different from the idealized CAD-derived geometry used in the \textit{Known-shape} setting. 

    \subsubsection{Surface Normal Estimation}
        Since the point clouds did not include normals, we estimated surface normals using Open3D (\texttt{estimate\_normals})~\cite{Open3D}.
        The normal directions were oriented consistently to point outward by flipping normals whose direction was inconsistent with the vector from the object CoM to each point.

    \begin{figure}[t]
        \centering
        \includegraphics[width=\columnwidth]{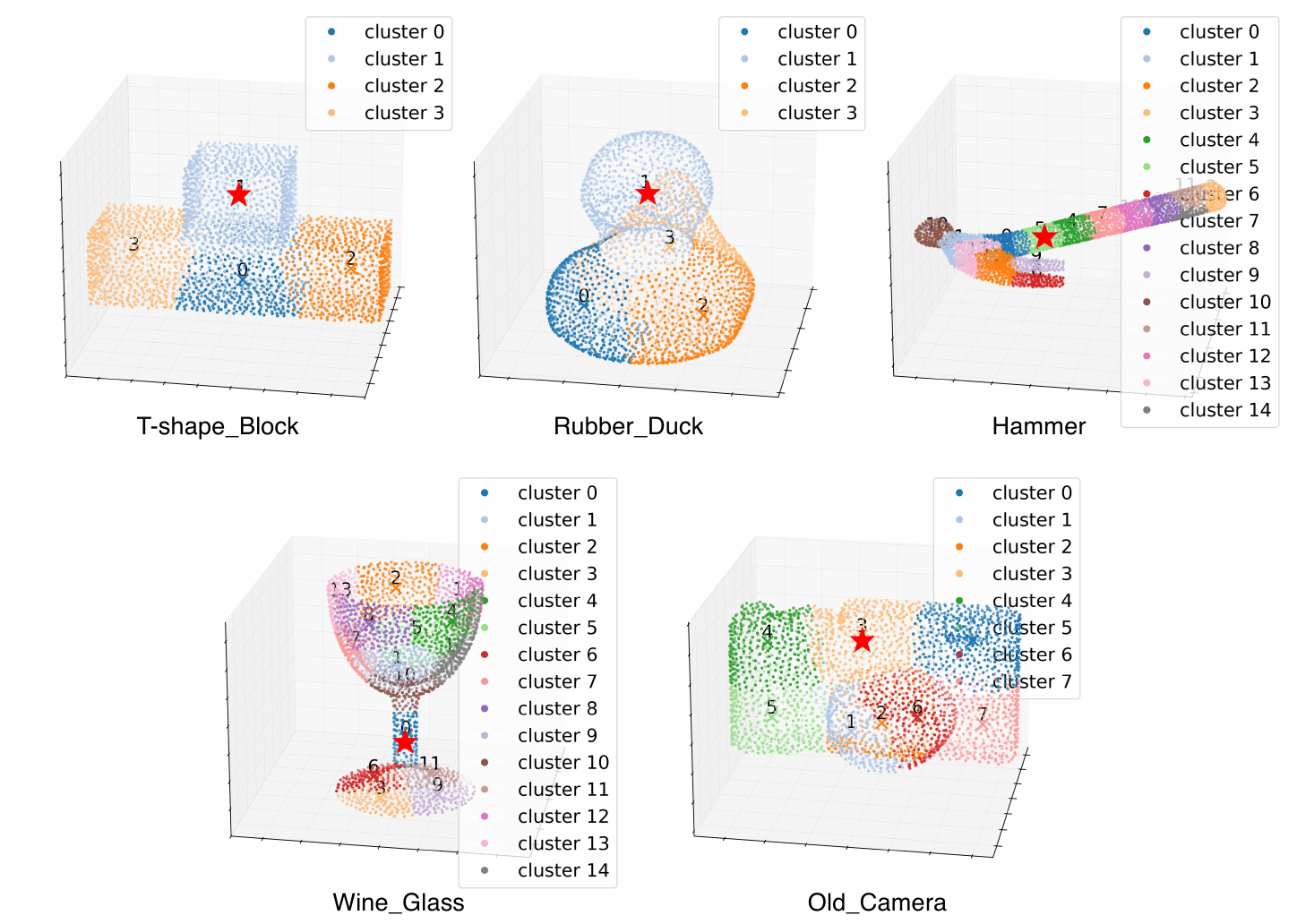}
        \caption{
            Initialization of grasp translation via $k$-means clustering. 
            For each object, we cluster the object point cloud into $k$ regions and treat the cluster centroids as candidate grasp positions. 
            The selected initial grasp position $\vt_0$ is indicated by the red marker.
        }
        \label{fig:kmeans_grasp_init}
    \end{figure}

    \subsubsection{Initialization of Grasp Pose}
        \label{sec:sim_exp_setup_grasp_pose}  
        
        The grasp translation $\vt_0$ was initialized using $k$-means clustering on the object point cloud: the cluster centroids were treated as candidate grasp positions, and one centroid was selected as the initial grasp position for optimization.
        Figure~\ref{fig:kmeans_grasp_init} illustrates this procedure. 
        The initial grasp orientation $\mR_0$ was manually designed for each object--robot pair to define a feasible approach direction consistent with the gripper geometry and workspace constraints.
        The initial gripper aperture $d_0$ was set to the maximum opening width of the gripper.

    \subsubsection{Comparative Methods}
        We compared three planners: (i) CMA-ES~\cite{CMA_ES}, a sampling-based optimizer, (ii) VISF~\cite{Fan_Case2018}, an iterative surface fitting baseline without CoM alignment, and (iii) DISF (ours), which integrates CoM alignment into disentangled iterative surface fitting.

    \subsection{Grasp Quality Evaluation}

        \subsubsection{Settings}
            We first evaluated grasp quality on the final optimized grasp pose using two metrics: the conventional geometric compatibility error and the CoM alignment error introduced in this study.
            This evaluation did not involve physics simulation or robot execution; instead, grasp poses were evaluated purely based on the object point cloud and a gripper surface model.
            Specifically, we used the Franka Hand\footnote{Franka Hand: $\texttt{\textcolor{blue}{https://franka.de/accessories}}$} gripper surface model to define the gripper surface and ran grasp planning for all 15 objects (5 Known-shape and 10 Observed-shape), reporting the resulting errors for each method.
            
            As an evaluation metric for geometric compatibility, we used the following weighted measure, which combines the surface distance defined in Eq.~\eqref{eq:Ep} and the contact normal misalignment defined in Eq.~\eqref{eq:En} with a scaling factor \( \alpha \): 
            \begin{equation} 
                \label{eq:E_geom}
                \begin{aligned}
                    E_{geom}(\vomega,\vt,\dd; \Sf, \So) = E_p(\vomega,\vt,\dd; \Sf, \So) + \alpha^2 E_{n}(\vomega; \Sf, \So). 
                \end{aligned}
            \end{equation} 
            The CoM misalignment was computed using the norm of the CoM difference between the gripper's and the object's canonical surfaces: 
            \begin{equation} 
                \label{eq:E_CoM}
                \begin{aligned}
                    E_{CoM}(\vomega,\vt,\dd; \Sf, \So) = \|\texttt{centroid}(\pO) - \texttt{centroid}(\pF^*) \|, 
                \end{aligned}
            \end{equation} 
            where $\pF^*$ is the canonical gripper surface transformed by the optimal grasping parameter $(\mR(\vomega^*), \vt^*, \dd^*)$. 

            In the experiments, we set the parameter $\alpha=0.1$, $\beta=0.85$, $d_0=0.091$, $d_{\mathrm{min}}=0.011$, $d_{\mathrm{max}}=0.091$, $\vv_0=[0, 1, 0]$, $\vn_{z0} = [0, 0, 1]$, $\Delta e = 1\mathrm{e}{-4}$. 
            The weight parameters such as $\alpha$ and $\beta$ were selected empirically based on preliminary experiments that yielded stable performance across the tested objects.
            We also used the predefined approach direction $\vn_{app}$ for each object. 

    \begin{figure}[thbp]
        \centering
        \includegraphics[width=0.95\columnwidth]{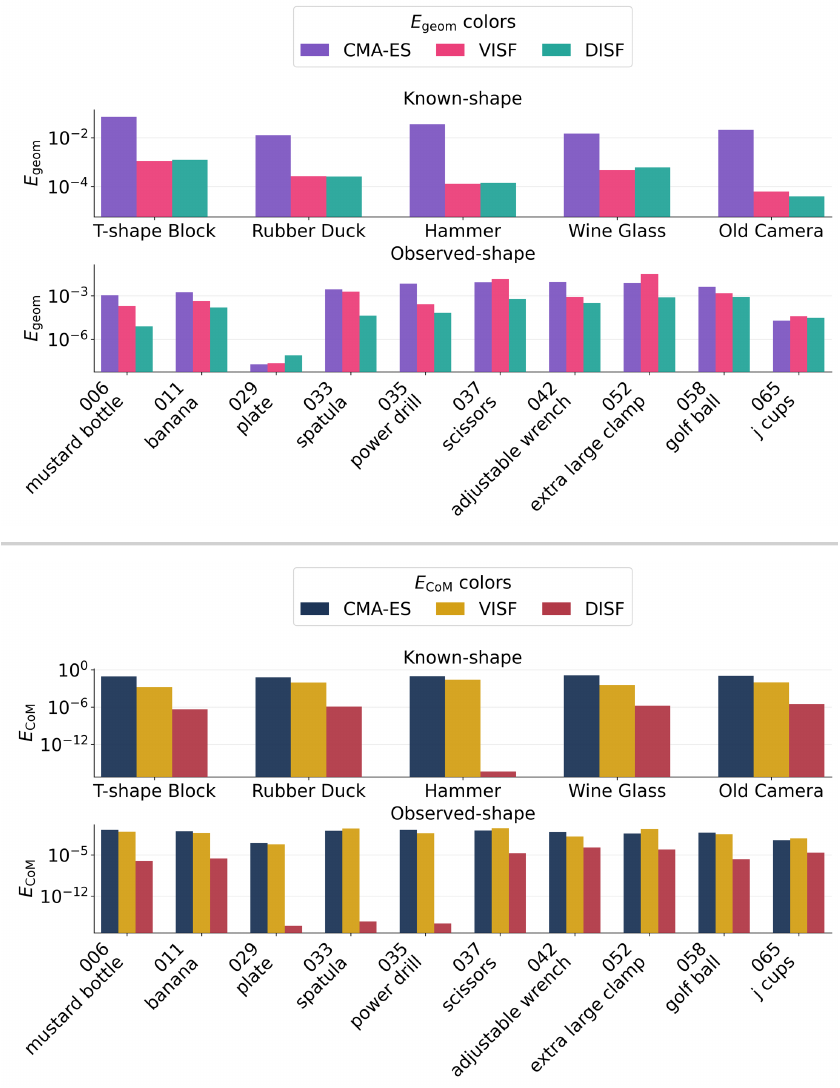}
        \caption{
            Grasp quality evaluation in simulation across the Known-shape and Observed-shape settings.
            The top figures report the geometric compatibility error $E_{geom}$, and the bottom figures report the CoM alignment error $E_{CoM}$, for the optimized grasp pose of each method.
            All values are shown on a logarithmic scale.
        }
        \label{fig:geom_com_errors}
    \end{figure}

    \subsubsection{Results}
        The results are shown in Fig.~\ref{fig:geom_com_errors}.
        The top row reports the geometric compatibility error $E_{{geom}}$, and the bottom row reports the CoM alignment error $E_{{CoM}}$.
        For each metric, we evaluated two shape-availability settings: the Known-shape setting (CAD-derived clean point clouds) and the Observed-shape setting (sensor-derived reconstructions).
        
        Regarding $E_{{geom}}$, VISF did not consistently achieve the lowest error despite explicitly optimizing geometric compatibility.
        A plausible explanation is that VISF simultaneously updates multiple pose parameters within a coupled optimization problem, where kinematic constraints prevent independent motion of the gripper contact surfaces.
        This coupling can increase sensitivity to local minima, especially when the object geometry is incomplete or noisy.
        
        In contrast, DISF maintained low $E_{{geom}}$ while consistently achieving the lowest $E_{{CoM}}$ across both settings.
        The improvement in $E_{{CoM}}$ was more pronounced in the Observed-shape setting, where noise and missing surfaces made it harder for the baselines to maintain CoM-consistent contact configurations.
        
        Overall, DISF preserved geometric compatibility while improving CoM alignment, which was expected to promote a more stable contact configuration and, consequently, higher grasp success in physics-based execution.

    \begin{figure}[t]
        \centering
        \includegraphics[width=\columnwidth]{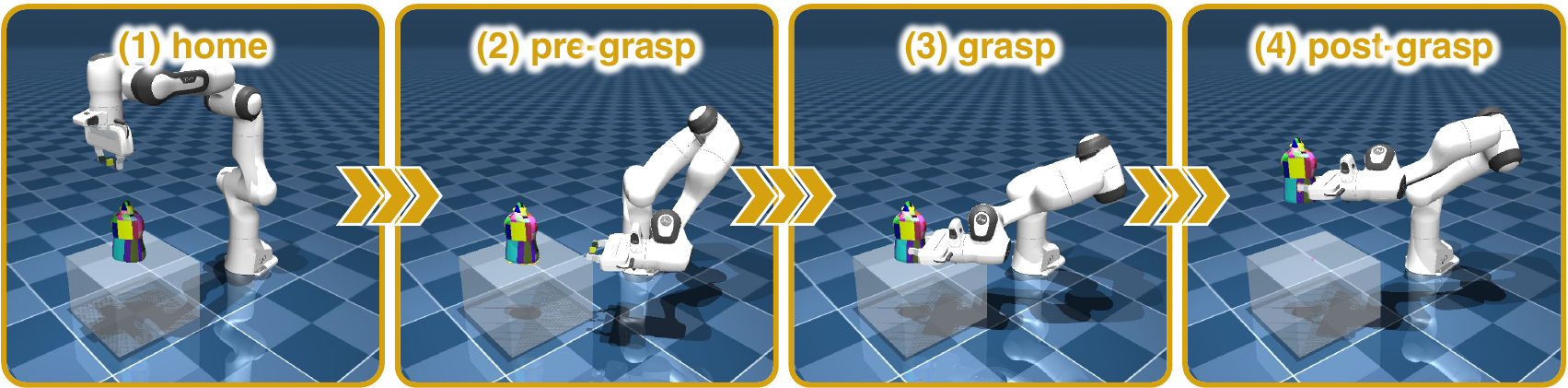}
        \caption{
            Grasp execution pipeline in MuJoCo used for success rate evaluation. 
            For each object, we executed the grasp planned from the optimal pose $(\mR(\vomega^*), \vt^*)$ with the optimal gripper aperture $\dd^*$.
            First, the robot moved its end-effector to a pre-grasp pose computed by offsetting the planned grasp pose along the approach direction.
            Next, from the pre-grasp pose, the robot approached the object to reach the final grasp pose and closed the gripper to the target aperture $\dd^*$.
            After grasping, the object was lifted vertically by $30$~cm and held stationary for $1$~s.
            Finally, grasp success was evaluated from the positional and orientational errors of the lifted object. 
        }        \label{fig:grasp_execution_pipeline}
    \end{figure}

    \subsection{Grasp Success Rate Evaluation}    
        \subsubsection{Settings}
            To evaluate the feasibility of the planned grasps, we executed them in the MuJoCo physics simulator~\cite{mujoco,menagerie2022github} and measured grasp success rates.

            Since DISF provided only the final grasp pose, we computed a pre-grasp pose and executed a simple approach trajectory to the final pose for grasp execution (Fig.~\ref{fig:grasp_execution_pipeline}).
            The pre-grasp pose definition and trajectory procedure are detailed in~\ref{appendix:pre-grasp_and_trajectory}.
            
            We evaluated success using the positional and orientational errors after lifting:
            \begin{subequations}
            \label{eq:grasp_evaluation_error}
            \begin{align}
            \mathbf{e}_{\textrm{pos}} &= \|\vp_{\textrm{target}} - \vp_{\textrm{lift}} \|, \\
            \mathbf{e}_{\textrm{ori}} &= 2 \cdot \arccos(|w|),
            \end{align}
            \end{subequations}
            where $\vp_{\textrm{target}}$ is the target position determined from the initial object position and the lifting height, $\vp_{\textrm{lift}}$ is the observed object position after execution, and $w$ is the scalar component of the quaternion representing the object's orientation.
            A trial was considered successful if both errors remained below predefined thresholds,
            $\eta_{\text{pos}}=0.06$~m and $\eta_{\text{ori}}=30^\circ$:
            \begin{equation}
            \label{eq:grasp_evaluation_condition}
            \text{Success} =
            \begin{cases}
            1, & \text{if } (\mathbf{e}_{\text{pos}} < \eta_{\text{pos}}) \land (\mathbf{e}_{\text{ori}} < \eta_{\text{ori}}), \\
            0, & \text{otherwise}.
            \end{cases}
            \end{equation}
            
            To mimic safety considerations in real-world execution, we additionally treated overly aggressive push-down motions as failures.
            During the approach/grasp phase in Fig.~\ref{fig:grasp_execution_pipeline}, if the object was pushed downward along the world $z$ axis by more than $0.01$~m from its initial height, we terminated the execution and counted the trial as a failure.

            To ensure reliable grasp execution in simulation, we applied an additional refinement to the fingertip displacement (i.e., a small closing bias) during execution.
            Details are provided in~\ref{appendix:fingertip_refine}.

    \begin{table}[t]
      \centering
      \caption{
        Grasp success on the Panda robot for the Known-shape and
        Observed-shape regimes. The Known-shape regime uses 3D CAD objects,
        while the Observed-shape regime uses YCB objects.
        The checkmark ($\greencheck$) indicates success, while the horizontal bar (-) indicates failure.  
        The bottom row reports the overall grasp success rate for each method across all objects of each setting and its average planning time. 
      }
      \label{tab:task_difficulty_panda}
      \small 
      \setlength{\tabcolsep}{4pt}
      \begin{tabular}{llccc}
        \toprule
        Setting & Object & CMA-ES & VISF & DISF (ours) \\
        \midrule
        \multirow{5}{*}{Known-shape}
         & \texttt{T-shape\_Block} & - & $\greencheck$ & $\greencheck$ \\
         & \texttt{Rubber\_Duck} & - & $\greencheck$ & $\greencheck$ \\
         & \texttt{Hammer} & - & $\greencheck$ & $\greencheck$ \\
         & \texttt{Wine\_Glass} & - & - & $\greencheck$ \\
         & \texttt{Old\_Camera} & - & $\greencheck$ & $\greencheck$ \\
        \midrule
        \multirow{10}{*}{Observed-shape}
         & \texttt{006\_mustard\_bottle} & - & - & $\greencheck$ \\
         & \texttt{011\_banana} & - & - & $\greencheck$ \\
         & \texttt{029\_plate} & - & - & - \\
         & \texttt{033\_spatula} & - & - & - \\
         & \texttt{035\_power\_drill} & - & - & $\greencheck$ \\
         & \texttt{037\_scissors} & - & - & $\greencheck$ \\
         & \texttt{042\_adjustable\_wrench} & - & - & $\greencheck$ \\
         & \texttt{052\_extra\_large\_clamp} & - & - & $\greencheck$ \\
         & \texttt{058\_golf\_ball} & - & - & $\greencheck$ \\
         & \texttt{065\text{-}j\_cups} & - & - & - \\
        \midrule
        \multicolumn{2}{l}{Success rate (Known-shape)} & 0/5 & 4/5 & \textbf{5/5} \\
        \multicolumn{2}{l}{Success rate (Observed-shape)} & 0/10 & 0/10 & \textbf{7/10} \\
        \multicolumn{2}{l}{Planning time [ms]} & 186.7 & \textbf{5.7} & 9.4 \\
        \bottomrule
      \end{tabular}
      \setlength{\tabcolsep}{6pt} 
    \end{table}

    \subsubsection{Results on the Panda robot}
        Table~\ref{tab:task_difficulty_panda} reports the grasp execution results on the Panda robot.
        DISF achieved the highest success rates in both settings, succeeding on all Known-shape objects (5/5) and on 7 out of 10 Observed-shape objects.
        In contrast, VISF succeeded on 4/5 objects in the Known-shape setting but failed on all Observed-shape objects (0/10),
        indicating that geometric-compatibility-based surface fitting alone was not robust to sensor-derived geometry with noise, missing surfaces and artifacts.
        CMA-ES failed on all objects in this evaluation. 
        
        Beyond success rate, Table~\ref{tab:task_difficulty_panda} also shows the computational efficiency.
        CMA-ES required 186.7~ms on average, whereas the surface-fitting methods were substantially faster (5.7~ms for VISF and 9.4~ms for DISF).
        This indicates that DISF improved grasp feasibility while retaining the practical runtime of iterative surface fitting.

    \begin{table}[t]
      \centering
        \caption{
        Average grasp success rate (\%) across robot--gripper platforms and evaluation regimes.
        We report results for three platforms (Panda+Franka Hand, UR5e+Robotiq HAND-E, iiwa+UMI gripper), each averaged over all objects in the corresponding setting.
        }
      \label{tab:robot_generalization}
      \small
      \begin{tabular}{llccc}
        \toprule
            Setting & Robot--gripper platform & CMA-ES & VISF & DISF (ours) \\
            \midrule
            \multirow{3}{*}{Known-shape}
             & Panda + Franka Hand & 0 & 80 & \textbf{100} \\
             & UR5e + HAND-E & 20 & 80 & \textbf{100} \\
             & iiwa + UMI gripper & 0 & 40 & \textbf{80} \\
            \cmidrule(lr){2-5}
             & \textbf{Average over platforms} & 7 & 67 & \textbf{93} \\
            \midrule
            \multirow{3}{*}{Observed-shape}
             & Panda + Franka Hand & 0 & 0 & \textbf{70} \\
             & UR5e + HAND-E & 30 & 40 & \textbf{60} \\
             & iiwa + UMI gripper & 10 & 60 & \textbf{80} \\
            \cmidrule(lr){2-5}
             & \textbf{Average over platforms} & 13 & 33 & \textbf{70} \\
            \midrule
            \multicolumn{2}{l}{\textbf{Total Average}} & 10 & 50 & \textbf{82} \\
        \bottomrule
      \end{tabular}
    \end{table}

    \begin{figure}[thbp]
        \centering
        \includegraphics[width=0.97\columnwidth]{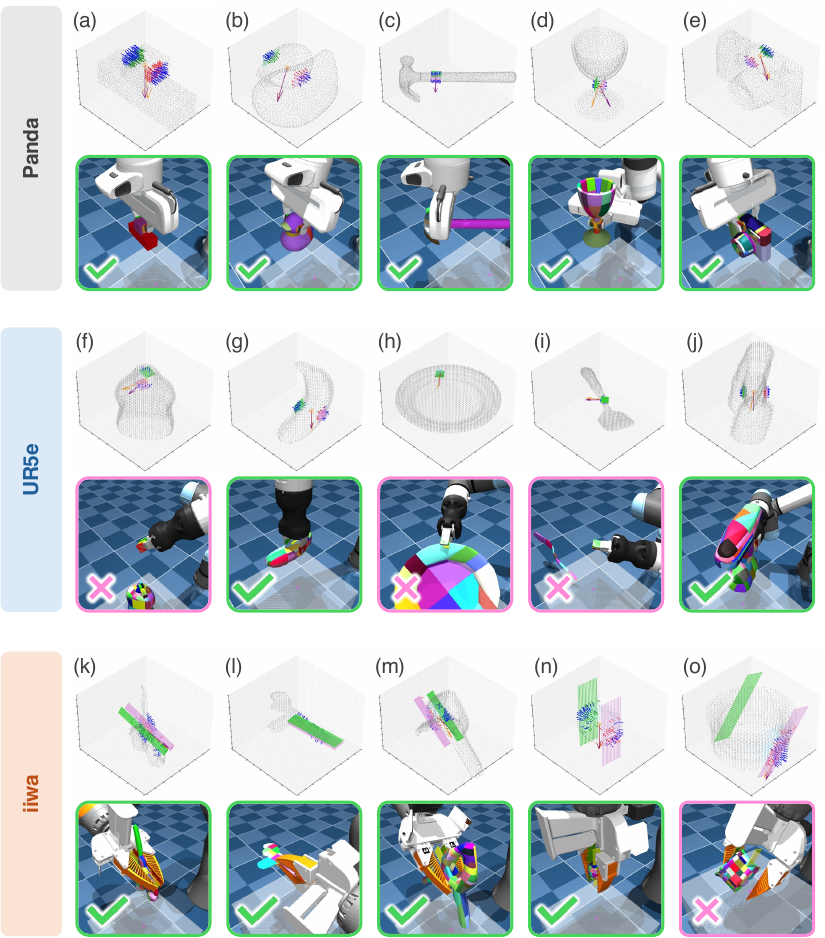}
        \caption{
            Examples of grasp executions planned by DISF across three robot--gripper platforms in Mujoco.
            From top to bottom: Panda, UR5e and iiwa. 
            For each object (a--o), the upper inset visualizes the planned grasp pose
            on the point cloud space where the visual elements include contact points (cyan), surface fitting points (blue), and fingertip surfaces (plum and lime green). 
            The orange arrow indicate the optimized z-axis direction of the gripper, while the purple arrow indicates a given approach direction $\vn_{app}$. 
            The lower image shows the corresponding execution outcome.
            Green frames with a checkmark indicate successful grasps, while magenta frames with a cross indicate failures.
            The object order matches Table~\ref{tab:task_difficulty_panda}: (a)--(e) Known-shape, (f)--(o) Observed-shape.
        }    
        \label{fig:grasp_results_examples}
    \end{figure}

    \subsubsection{Cross-platform Evaluation (Generalization across Grippers)}
        To evaluate generality across diverse grippers, Table~\ref{tab:robot_generalization} summarizes the average grasp success rates on three representative robot--gripper setups (see Fig.~\ref{fig:robot_platforms}).
        While the manipulator models also differ across these platforms, our primary interest here is robustness to gripper-geometry variability, i.e., whether DISF transfers across different grippers without redesigning the planner.
        Table~\ref{tab:gripper_specs} summarizes the fingertip surface size and aperture range of each gripper.
        
        DISF consistently outperformed the baselines across all gripper setups, achieving 93\% average success in the Known-shape setting and 70\% in the Observed-shape setting,
        resulting in an overall average of 82\%.
        Notably, the improvement of DISF over VISF was larger in the Observed-shape setting
        (+37 points: 70\% vs.\ 33\%) than in the Known-shape setting (+26 points: 93\% vs.\ 67\%), suggesting that explicitly promoting CoM alignment in surface fitting can improve robustness when object geometry is imperfect.
        
        Fig.~\ref{fig:grasp_results_examples} visualizes representative execution outcomes across platforms. 
        Successful cases show that DISF yields feasible grasps that remain stable during lifting,
        whereas the failures reveal limitations inherent to surface-fitting-based planning:
        (i) contact interactions (e.g., friction and local contact modeling) are not explicitly optimized, and
        (ii) CoM alignment is computed from the available surface representation, which may differ from the true physical CoM.
        These limitations are particularly evident for thin or highly asymmetric objects (e.g., \texttt{029\_plate} and \texttt{033\_spatula}),
        where incomplete contact or rotational slip can occur even when the planned grasp appears geometrically plausible.
        
        Overall, the results demonstrate that DISF improves grasp success by enhancing contact stability via CoM alignment,
        while preserving the efficiency of iterative surface fitting.

\section{Real-world Experiments}
    We conducted real-world grasp executions to evaluate whether grasps planned by our surface fitting algorithm transfer to a physical setup under real-world uncertainties, including variations in friction and contact dynamics, calibration errors, and imperfect object observations (e.g., partial views and sensor noise).
    We evaluated three planners (CMA-ES, VISF, and DISF) on a UR3e manipulator equipped with a Robotiq Hand-E gripper, and reported grasp success rate as the primary metric.
    To mirror the evaluation settings used in simulation, we performed experiments under both a \emph{Known-shape} setting and an \emph{Observed-shape} setting.
    In the Known-shape setting, grasps were planned from CAD-derived object point clouds (3D models) for a set of 3D-printed objects.
    In the Observed-shape setting, grasps were planned directly from camera-observed point clouds captured by depth sensors.

    \begin{figure}[t]
        \centering
        \includegraphics[width=\columnwidth]{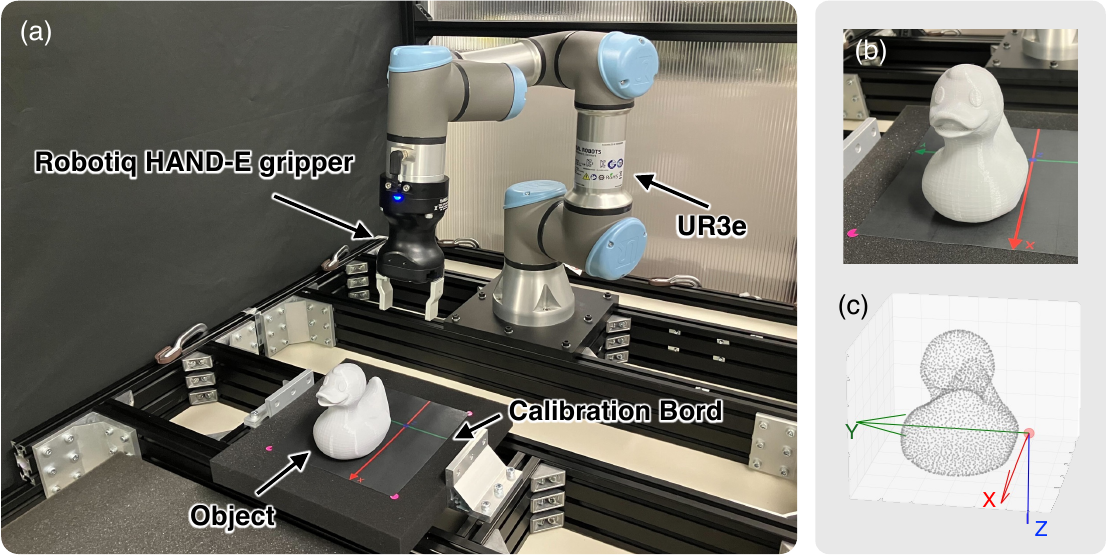}
        \caption{
            Real-world experimental setup. 
            (a) A UR3e manipulator equipped with a Robotiq Hand-E parallel gripper.
            Objects were placed on a printed calibration board (grid and reference axes) to define a consistent planning frame in the workspace.
            (b) Example object placement on the calibration board.
            (c) The grasp-planning coordinate frame used throughout the real-world experiments.
            All planned grasp poses were expressed in this frame, and the robot task-space reference was calibrated accordingly.
        }
        \label{fig:real_setup}
    \end{figure}

\subsection{Common Setup}
    This section describes the experimental components shared by both the Known-shape and Observed-shape settings.
    Unless otherwise stated, we used the same hardware, calibration procedure, grasp initialization, execution pipeline, and evaluation protocol across the two settings.

    \subsubsection{Hardware Platform}
        All real-world experiments were conducted using a UR3e manipulator equipped with a Robotiq Hand-E parallel-jaw gripper.
        The robot executed grasp motions in a tabletop workspace, as illustrated in Fig.~\ref{fig:real_setup}.

    \subsubsection{Workspace and Frame Calibration}
        To align the real-world workspace with the grasp-planning coordinate system, we placed a printed calibration board on the work surface (Fig.~\ref{fig:real_setup} (a)).
        The board was a planar target showing a 30-mm grid together with Cartesian axes.
        We then adjusted the robot task-space reference such that, at a predefined nominal pose in the planning frame
        (translation $\mathbf{t}=\mathbf{0}$ and rotation $\mathbf{R}=\mathbf{I}$),
        the gripper center was aligned with the origin of the calibration board.         Throughout the experiments, objects were placed on the board so that their nominal placement matched the grasp-planning space.

    \subsubsection{Initialization and Execution Protocol}
        We followed the same initialization and execution procedure as in simulation described in Sec.~\ref{sec:sim_exp_setup_grasp_pose}. 
        The grasp translation $\mathbf{t}_0$ was initialized via k-means clustering on the object point cloud, while the rotation $\mathbf{R}_0$ was manually specified to yield a feasible approach direction for each object, and the initial gripper aperture $d_0$ was set to the gripper's maximum opening width (e.g. $d = 0.05$ [m]).
        Because the point-cloud characteristics differ between Known-shape and Observed-shape settings (e.g., point density and occlusions), the k-means configuration and selected centroid were tuned per setting.
        We also adopted the same grasp execution procedures in the experiments as in simulation.

    \subsubsection{Evaluation Protocol}
    \label{sec:eval_protocol}
        In real-world setup, it is difficult to obtain the object's 6-DoF pose with the same accuracy as in simulation. Therefore, directly judging grasp success based on complete pose-state information (e.g., the true post-lift pose change) is not feasible. 
        To address this, we determine grasp success using a post-grasp close-probing procedure, where an additional gripper-closing command is issued after the post-grasp motion. Intuitively, if the grasp has not been established, the object does not constrain the gripper and the fingers can continue closing; consequently, the gripper aperture decreases to near zero. In contrast, if the grasp is established, the object thickness prevents further closing, and the gripper aperture remains above a certain value. 
        
        Let $d$ [mm] denote the gripper aperture measured after the close-probing step, and let $\xi$ [mm] be a threshold that represents the minimum aperture indicating that an object remains between the fingers. The protocol is as follows: (1) after executing the grasp and the post-grasp lifting motion, we send an additional closing command to the gripper; (2) after waiting for $\delta t$ [sec], we read the gripper aperture $d$ and declare the grasp successful if $d > \xi$. 
        This criterion avoids sensitivity to force-sensor drift and noise while keeping the implementation simple. In our experiments, we set $\delta t = 1.0$ and $\xi = 3.0$.
        
        Any execution that was aborted by the controller (e.g., due to collisions or kinematically unsafe was counted as a failure.

    \begin{figure}[t]
        \centering
        \includegraphics[width=\columnwidth]{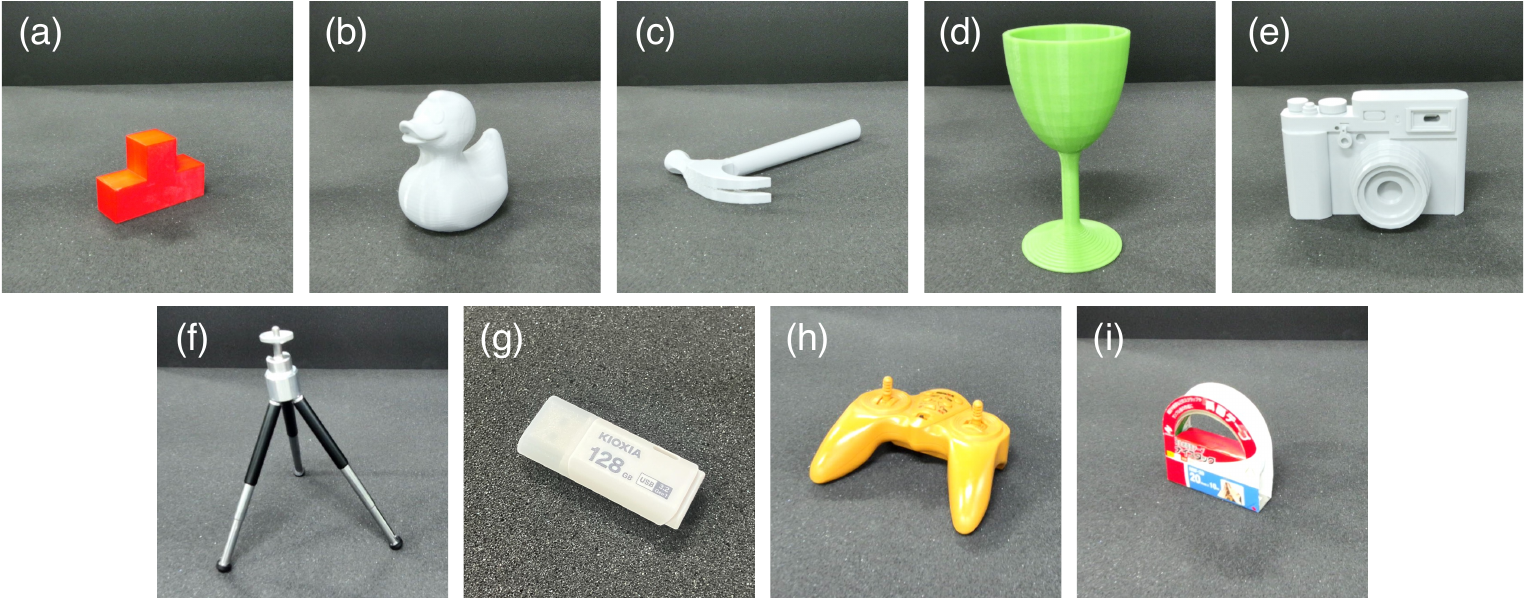}
        \caption{
            Objects used in the real-world evaluation.
            Objects (a)–(e) were 3D-printed using PLA filament and were used in both the Known-shape and Observed-shape settings to enable a direct comparison, while (f)–(i) were additional objects evaluated only in the Observed-shape setting. 
            (a) \texttt{T-shape\_Block}, 
            (b) \texttt{Rubber\_Duck}, 
            (c) \texttt{Hammer}, 
            (d) \texttt{Wine\_Glass}, 
            (e) \texttt{Old\_Camera}, 
            (f) \texttt{Tripod}, 
            (g) \texttt{USB}, 
            (h) \texttt{Controller}, and 
            (i) \texttt{Tape}. 
        }
        \label{fig:real_objects}
    \end{figure}

\subsection{Known-shape Setting}
\label{sec:real_known}
    
    \subsubsection{Setup}
    \label{sec:real_known_setup}
        In the Known-shape setting, we executed grasps planned from the same object models used in simulation. 
        In the real setup, each object was placed on the calibration board so that its nominal pose matched the predefined grasp-planning coordinate frame, thereby ensuring consistency between the planned grasp pose and the physical placement.
        We evaluated the five objects (a)--(e) in Fig.~\ref{fig:real_objects} to enable a direct comparison with the Observed-shape setting.

        \subhead{Grasp parameter setting}
        Since the object geometry is available as a clean CAD model in this setting, we did not re-plan grasp pose parameters. 
        Instead, we reuse the grasp poses obtained in simulation under the Known-shape condition and directly execute them on the physical robot. 
        This setting therefore evaluates whether grasps planned under accurate shape information transfer reliably to real execution.

    \subsubsection{Results}
    \label{sec:real_known_results}
        
        Table~\ref{tab:real_grasp_success} (Known-shape block) summarizes real execution when clean object geometry is available and the grasp poses are directly transferred from simulation.
        Under this setting, DISF succeeds on all trials, and VISF achieves the same outcome pattern as in simulation, failing only on \texttt{Wine\_Glass}. 
        This indicates that, with clean object geometry, the planned grasps transfer to real execution with identical success rates for both methods.
        CMA-ES fails on all trials, consistent with its behavior in simulation, indicating that direct black-box search over grasp parameters is not reliable.

        \begin{table}[t]
          \centering
          \caption{
            Real-world grasp success results under the Known-shape and Observed-shape settings.
            The Observed-shape setting is reported in two blocks: (i) the same objects as the Known-shape setting (enabling a direct Known vs.\ Observed comparison), and (ii) additional objects used only in the Observed-shape setting.
            The checkmark ($\greencheck$) indicates success, while the horizontal bar (-) indicates failure.
          }
          \label{tab:real_grasp_success}
          \setlength{\tabcolsep}{4pt}
          \footnotesize
          \begin{tabular}{llccc}
            \toprule
            Setting & Object & CMA-ES & VISF & DISF (ours) \\
            \midrule
            \multirow{5}{*}{\makecell[l]{Known-shape\\(real execution)}}
             & \texttt{T-shape\_Block} & - & $\greencheck$ & $\greencheck$ \\
             & \texttt{Rubber\_Duck}   & - & $\greencheck$ & $\greencheck$ \\
             & \texttt{Hammer}        & - & $\greencheck$ & $\greencheck$ \\
             & \texttt{Wine\_Glass}   & - & -             & $\greencheck$ \\
             & \texttt{Old\_Camera}   & - & $\greencheck$ & $\greencheck$ \\
            \midrule
        
            \multirow{5}{*}{\makecell[l]{Observed-shape\\(same objects)}}
             & \texttt{T-shape\_Block} & - & $\greencheck$ & $\greencheck$ \\
             & \texttt{Rubber\_Duck}   & - & $\greencheck$ & $\greencheck$ \\
             & \texttt{Hammer}        & - & -             & $\greencheck$ \\
             & \texttt{Wine\_Glass}   & - & -             & $\greencheck$ \\
             & \texttt{Old\_Camera}   & - & -             & $\greencheck$ \\
            \midrule
        
            \multirow{4}{*}{\makecell[l]{Observed-shape\\(additional objects)}}
             & \texttt{Tripod}     & - & - & $\greencheck$ \\
             & \texttt{USB}        & - & - & $\greencheck$ \\
             & \texttt{Controller} & - & - & -             \\
             & \texttt{Tape}       & - & - & $\greencheck$ \\
            \midrule
        
            \multicolumn{2}{l}{Success rate (Known-shape)} & 0/5 & 4/5 & \textbf{5/5} \\
            \multicolumn{2}{l}{Success rate (Observed-shape, same objects)} & 0/5 & 2/5 & \textbf{5/5} \\
            \multicolumn{2}{l}{Success rate (Observed-shape, additional objects)} & 0/4 & 0/4 & \textbf{3/4} \\
            \multicolumn{2}{l}{Success rate (Observed-shape, overall)} & 0/9 & 2/9 & \textbf{8/9} \\
            \bottomrule
          \end{tabular}
          \setlength{\tabcolsep}{6pt}
        \end{table}

    \begin{figure}[t]
        \centering
        \includegraphics[width=\columnwidth]{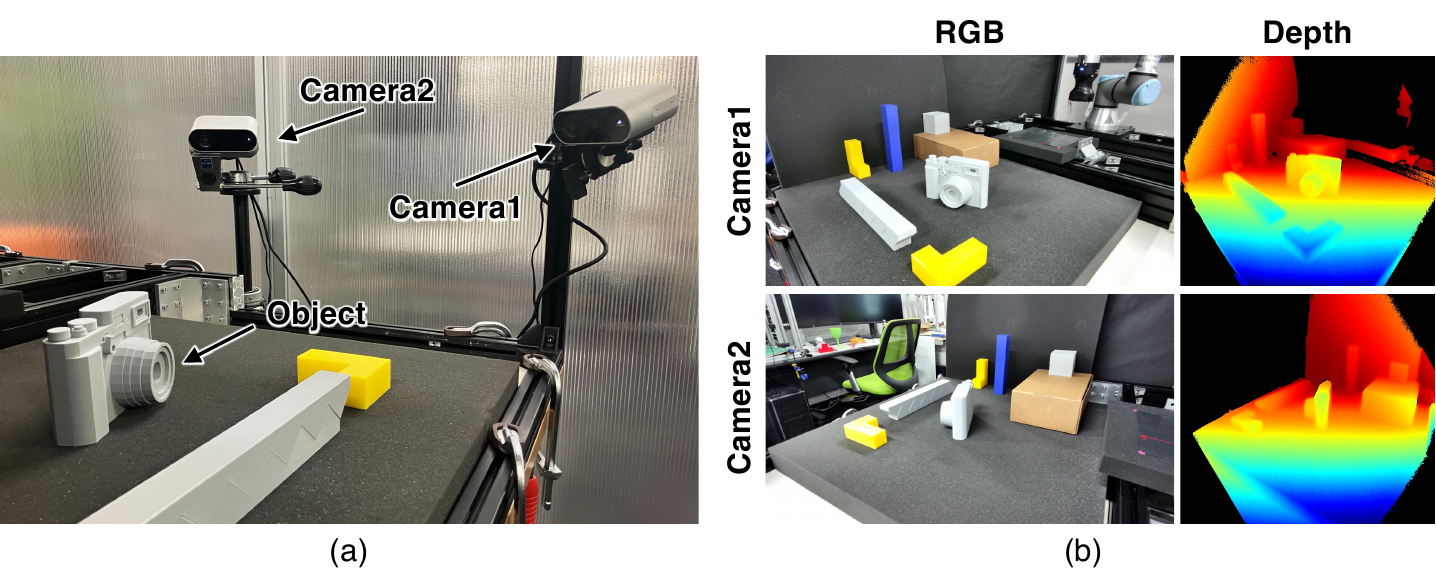} 
        \caption{
            Two-view depth sensing setup using two Orbbec Femto Bolt sensors (Camera1: left view, Camera2: right view). 
            Prior to point cloud generation and multi-view registration, depth values are clipped to camera-specific valid ranges (Camera1: 200–700 mm; Camera2: 200–1000 mm) to suppress invalid readings. 
            Multiple static landmark objects are placed in the scene to improve overlap and stabilize ICP-based cross-view registration.
        }
        \label{fig:camera_setup}
    \end{figure}

    \begin{figure}[thbp]
        \centering
        \includegraphics[width=\columnwidth]{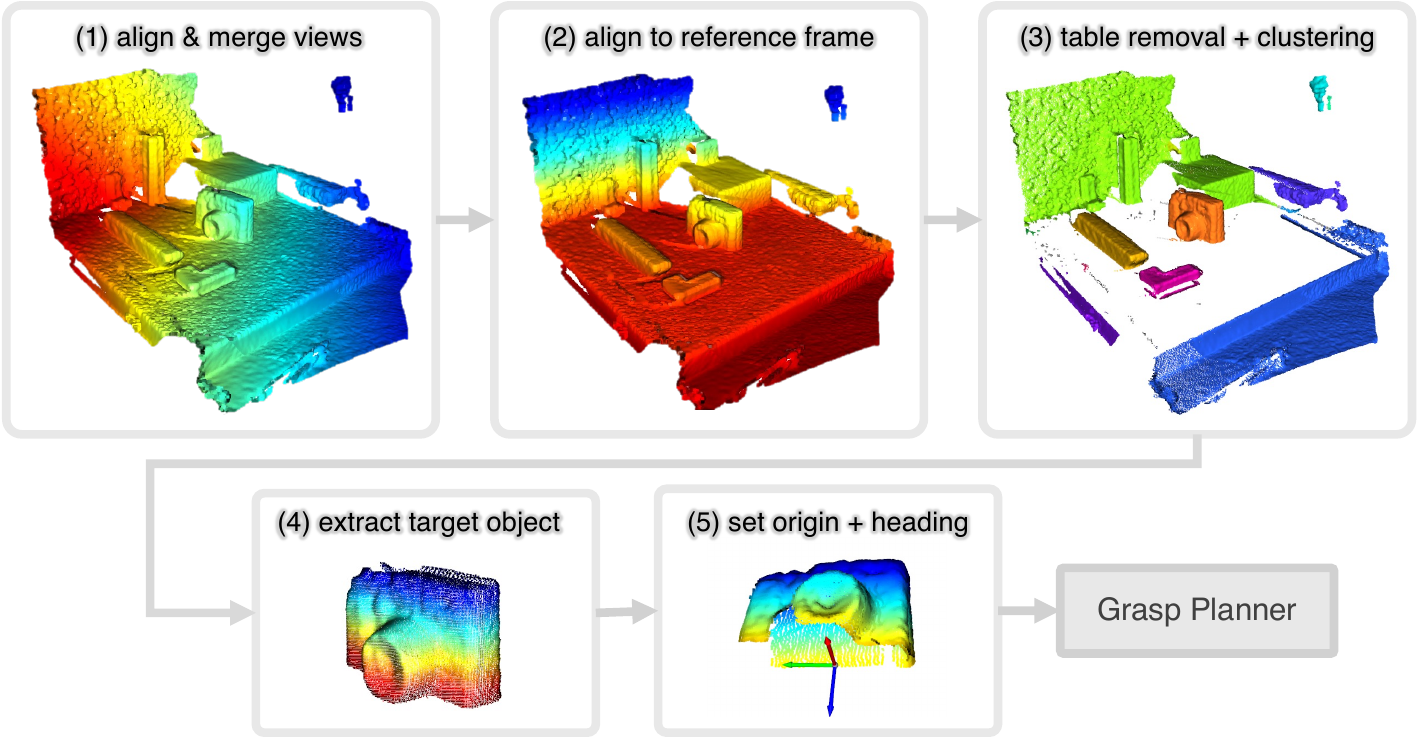}
        \caption{
            Preprocessing pipeline for converting two-view depth observations into a planner-ready object point cloud. 
            (1) Register and merge the two view point clouds into a single scene cloud. 
            (2) Align the scene with the gripper approach axis (z-axis). 
            (3) Remove the support plane (table) and cluster remaining points into object candidates. 
            (4) Extract and denoise the target object cluster. 
            (5) Set gripper-frame origin and heading. 
        }   
        \label{fig:pc_preprocess}
    \end{figure}

\subsection{Observed-shape Setting}
\label{sec:real_observed}

    \subsubsection{Camera Setup}
    \label{sec:real_observed_setup}

         To construct observed object point clouds for grasp planning, we use two depth sensors (Orbbec Femto Bolt) placed around the workspace to capture complementary views of the scene (Fig.~\ref{fig:camera_setup}). 
         We refer to the sensors as Camera1 (left view) and Camera2 (right view).

        \subhead{Depth range clipping}
        To improve the stability of subsequent point cloud registration, we clip raw depth measurements to camera-specific valid ranges and discard values outside these intervals.
        This reduces far-range noise and background clutter while retaining the geometry around the target object.

        \subhead{Landmarks for robust registration}
        Because multi-view registration can be ill-conditioned when the scene contains limited geometric variation, we intentionally place multiple landmark objects in the scene in addition to the target object to be grasped.
        These landmarks introduce distinctive geometric structure, improving the robustness of the global registration stage (RANSAC-based) and providing a reliable initialization for subsequent local refinement (ICP-based), resulting in more stable alignment between the two camera point clouds.

    \subsubsection{Point Cloud Preprocessing}
        Starting from the merged two-view scene point cloud, we apply a deterministic preprocessing pipeline to obtain a single object point cloud with a consistent coordinate frame suitable for the grasp planner (Fig.~\ref{fig:pc_preprocess}). 
        The pipeline standardizes the workspace orientation, removes the support surface, isolates the target object from clutter, and finally defines an object-local frame (origin and heading) used by the planner. We will describe the detail of each stage in the below. 
    
        \subhead{(1) Align and merge views}
        We first combine the two depth-derived point clouds captured from Camera1 (left view) and Camera2 (right view) into a single scene point cloud. 
        We estimate a rigid transformation that aligns the left-view point cloud to the right-view point cloud. 
        For robustness, we optionally remove sparse outliers from both views using statistical outlier removal. 
        We then compute a coarse initial alignment using a global registration procedure based on geometric feature matching (RANSAC~\cite{RANSAC} with FPFH descriptors~\cite{FPFH}) on voxel-downsampled point clouds. 
        Starting from this initialization, we refine the alignment with point-to-plane ICP~\cite{ICP_point2plane_2001} using estimated surface normals. 
        Finally, we transform the left-view point cloud into the right-view coordinate frame and merge the two point sets to obtain a unified scene point cloud used in subsequent steps.

        \subhead{(2) Align the scene with the gripper approach axis}
        To make the downstream processing consistent with the planner convention, we re-orient the merged scene so that the dominant support plane (the tabletop) provides a stable reference for the vertical axis. Concretely, we detect the support plane and rotate the scene such that the plane normal becomes aligned with the gripper approach axis (the gripper z-axis). We additionally shift the scene to use the tabletop as a consistent height reference.

        \subhead{(3) Table removal and clustering into object candidates}
        After alignment, we remove the support plane to isolate points belonging to objects placed on the table. We then cluster the remaining points based on spatial proximity to form object candidates. This step yields a set of clusters corresponding to individual objects (and potential small noise clusters), which are used for target selection.
        
        \subhead{(4) Extract and denoise the target object}
        We select the cluster corresponding to the target object by specifying a cluster ID. The extracted cluster is further cleaned by removing residual outliers and small disconnected fragments, producing a compact and denoised target-object point cloud suitable for grasp planning.

\begin{figure}[t]
        \centering
        \includegraphics[width=\columnwidth]{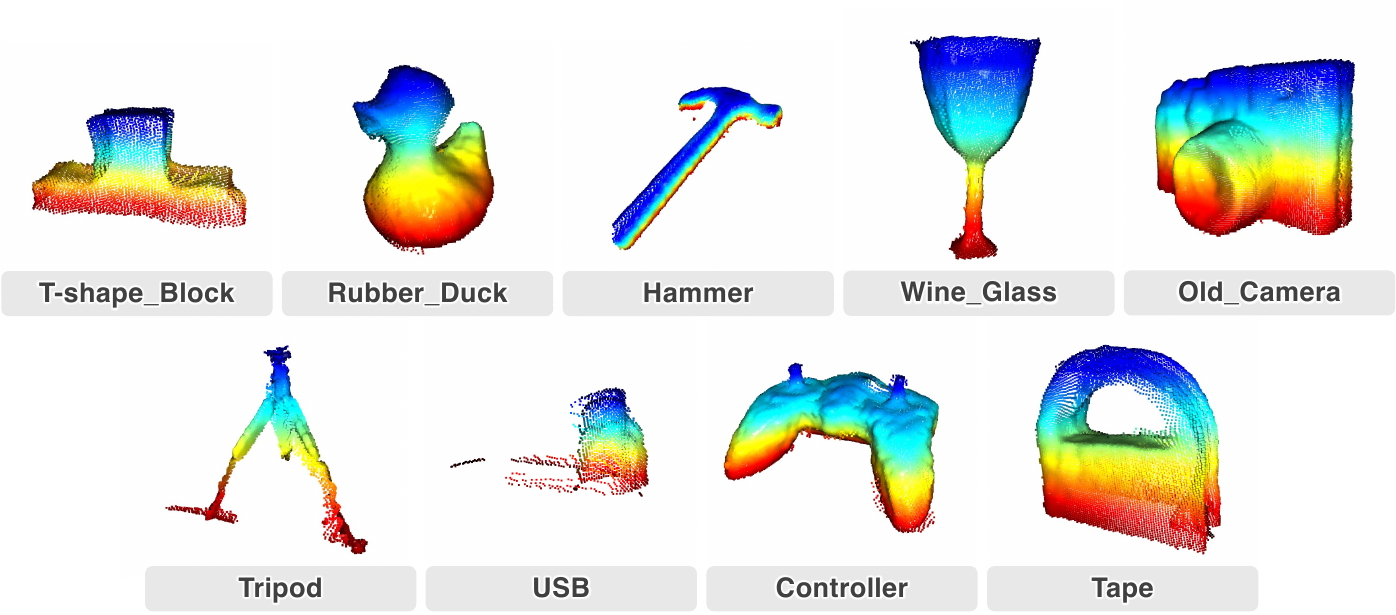} 
        \caption{
            Observed object point clouds after preprocessing.
        }
        \label{fig:observed_object_gallery}
    \end{figure}

        \subhead{(5) Specify object position and heading in gripper frame}
        Finally, to parameterize the object pose in the coordinate convention used by the planner, we assign two reference points on the extracted target. 
        The first point is chosen as an anchor and is treated as the origin of the gripper frame, i.e., the point that corresponds to $(0, 0, 0)$ in the gripper frame. 
        The second point defines a heading direction relative to the anchor and is used to specify the object's in-plane orientation (yaw). 
        With this anchor-and-heading convention, the target point cloud can be expressed consistently in the gripper frame, enabling grasp planning. The actual preprocessed object point clouds are shown in Fig.~\ref{fig:observed_object_gallery}.

    \subsubsection{Results}
    \label{sec:real_observed_results}
\begin{figure}[thbp]
        \centering
        \includegraphics[width=\columnwidth]{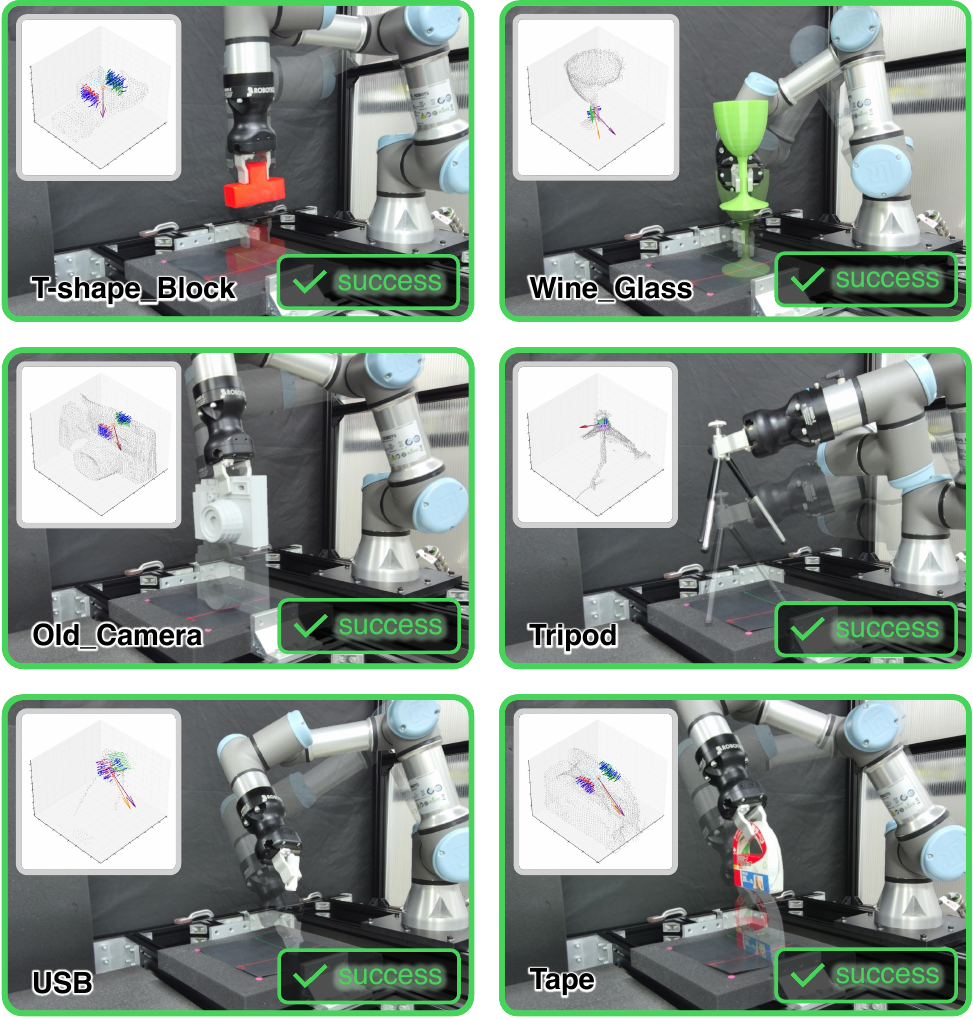}
        \caption{
            Successful real-world grasps planned by DISF under the Observed-shape setting. Each figure shows a different object grasped on the UR3e, including both objects shared with the Known-shape setting and additional objects evaluated only in the Observed-shape setting. 
            The inset in the upper-left of each figure visualizes the planner input: the reconstructed object point cloud in the gripper-aligned frame with the corresponding planned grasp pose.
        } 
        \label{fig:real_success_observed}
    \end{figure}

        Table~\ref{tab:real_grasp_success} (Observed-shape blocks) reports the grasp performance when grasps are planned from observed point clouds. 
        The performance gap between DISF and VISF widens markedly under observed geometry. 
        On the same objects as in the Known-shape evaluation, DISF achieves an identical success rate to the Known-shape setting, while VISF degrades. 
        CMA-ES again fails across all Observed-shape trials. 
        DISF further generalizes to additional everyday objects introduced only in the Observed-shape setting, demonstrating robustness to partial and noisy geometric observations.
        Examples of successful grasp are shown in Fig.~\ref{fig:real_success_observed}.

\subsection{Failure Mode Analysis}

    To better interpret the real-world outcomes under the Observed-shape setting, we analyze failures by categorizing each trial into three observable modes: \textit{contact-induced abort}, \textit{misgrasp}, and \textit{success}. 
    A \textit{contact-induced abort} occurs when execution is terminated by the robot's safety mechanisms due to unintended contacts during the pre-grasp or grasp phase in the execution. 
    A \textit{misgrasp} denotes trials where the planned motion is executed without safety stops, yet the grasp is not established at closure (e.g., the fingers miss the intended contact region or fail to securely grasp the object). 
    The remaining trials that satisfy the evaluation protocol described in Section~\ref{sec:eval_protocol} are counted as \textit{success}.
    
\begin{figure}[t]
        \centering
        \includegraphics[width=\columnwidth]{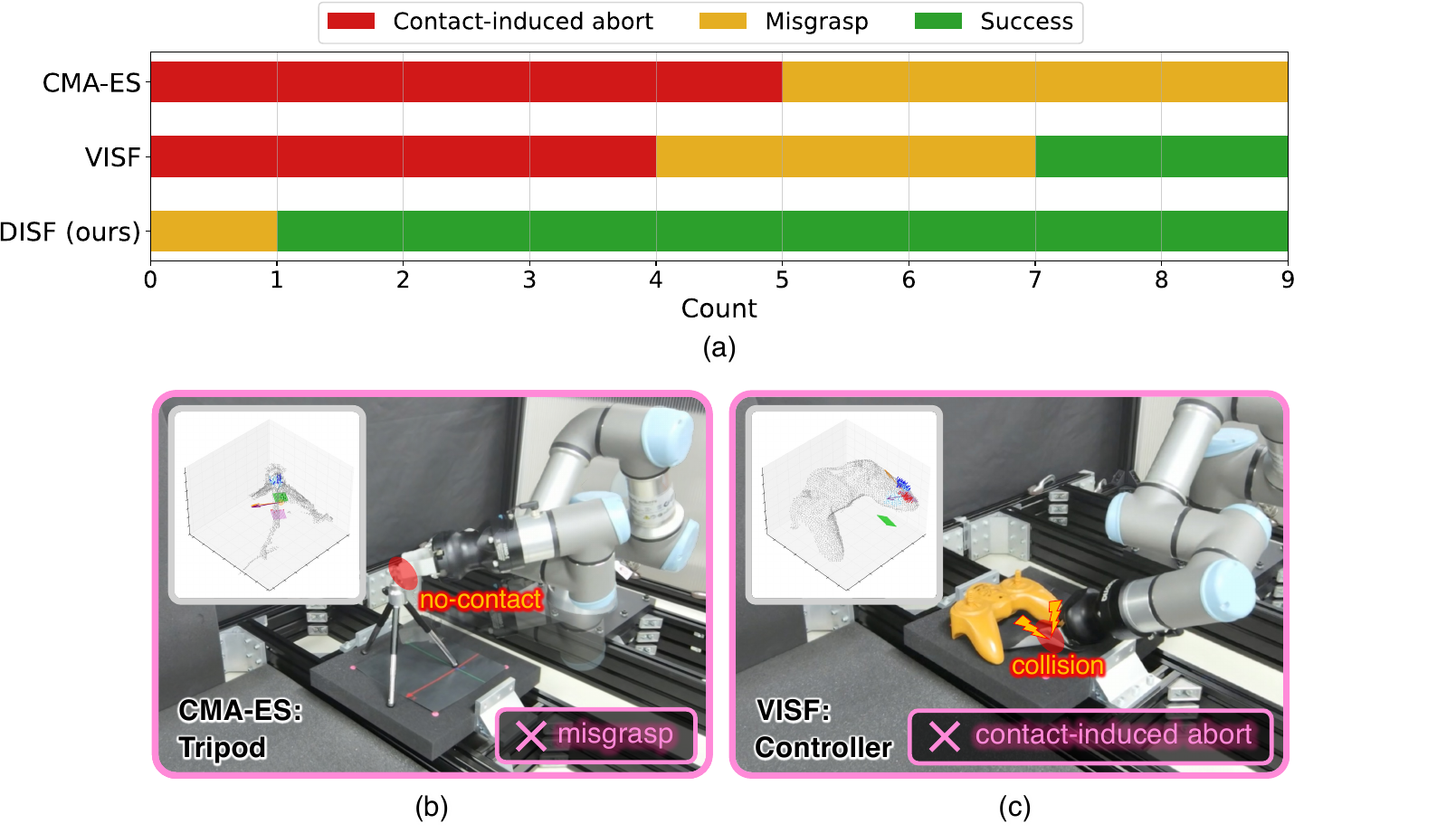}
        \caption{
            Failure-mode analysis in real-world grasp execution under the Observed-shape setting. (a) Failure-mode breakdown over 9 trials per method, categorized as \textit{contact-induced abort}, \textit{misgrasp}, or \textit{success} according to the evaluation protocol in Sec.~\ref{sec:eval_protocol}. (b--c) Representative failure cases illustrating the two dominant failure modes for the baselines: (b) a \textit{misgrasp} where the motion completes but the gripper fails to establish contact with the object (CMA-ES on Tripod), and (c) a \textit{contact-induced abort} triggered by unintended collision during approach, leading to an automatic safety stop (VISF on Controller). Insets visualize the planner input point cloud and the planned grasp pose for each trial.
        }
       \label{fig:failure_mode_breakdwon}
    \end{figure}

    The resulting failure-mode breakdown is shown in Fig.~\ref{fig:failure_mode_breakdwon} across the methods. 
    CMA-ES exhibits no successful trials; its outcomes are dominated by \textit{contact-induced aborts} and \textit{misgrasps}, indicating that black-box search often produces low-feasibility grasp plans under observed geometry.
    VISF achieves a limited number of \textit{successes}, but still shows a large portion of \textit{contact-induced aborts} and \textit{misgrasps}, suggesting sensitivity to observation imperfections that can perturb the planned approach and contact configuration.
    In contrast, DISF converts most trials into \textit{successes}, with only a single \textit{misgrasp} and no \textit{contact-induced abort} in our experiments. 
    This indicates that enforcing the CoM alignment in the surface fitting-based grasp planning algorithm yields safer and more reliable grasps when the object geometry is provided as reconstructed point clouds from camera observations.
    
    Overall, this failure-mode breakdown explains the pronounced performance gap observed in Table~\ref{tab:real_grasp_success} under observed geometry: baseline methods frequently fail either by triggering safety stops or by producing grasps that execute but do not establish stable contact, whereas DISF substantially reduces both failure modes.

\section{Discussion}
\label{sec:discussion}
We discuss several limitations of DISF and outline directions for future work toward more reliable and practical grasp planning.

\subsection{Limited physical/contact and kinematics modeling during planning}
    While the proposed framework optimizes geometric compatibility and contact stability through CoM alignment, it does not explicitly model friction, compliance, or contact forces inside the grasp planning.
    As a result, failures can still occur when a geometrically plausible contact configuration becomes unstable during execution (e.g., incomplete contact or rotational slip).
    Moreover, the lack of contact modeling also makes it difficult to reliably detect whether the planned grasp induces physically infeasible penetration between the gripper and the object, and thus collision-free execution is not guaranteed.
    Bridging this gap may require tighter coupling with contact-aware constraints~\cite{Kaiyu2016Hierarchical,Fan_RAL_2019,Fan_Sensors_2024}, or closed-loop execution with force/tactile feedback and online re-planning.

    Additionally, DISF does not explicitly incorporate robot kinematic reachability as a planning constraint (except indirectly through the approach-direction consistency term);
    therefore, a grasp pose that is feasible in the gripper--object grasp space can still fail in practice when the manipulator cannot reach the pose without violating joint limits or encountering collisions.
    Incorporating kinematics-aware constraints (e.g., IK feasibility and motion-planning constraints) is thus an important direction for improving real-world reliability~\cite{Kaiyu2016Hierarchical,Lie_ICRA_2023,Weiyu2023StructDiffusion}.

\subsection{Heuristic selection of correspondence pairs}
    To execute the proposed framework, we assume that reliable correspondence pairs between the object surface and the gripper surface can be obtained.
    However, the current implementation does not explicitly provide a dedicated pipeline for estimating such correspondences.
    Instead, we compute correspondences using distance- and rotation-based filtering with manually tuned heuristic thresholds to select valid pairs from a large set of candidate pairs.
    While this procedure is sufficient as an initial step to validate the core concept on relatively simple (often convex) objects, where many correspondences can be found for a wide range of grasp poses, it limits scalability to more complex geometries and tasks.
    In particular, for non-convex or thin objects (e.g., plates, spatulas, or narrow structures such as the rim/edge of a mug), we occasionally obtain correspondences only on one side (e.g., from a single fingertip region), which can lead to unstable or biased grasp optimization.
    Therefore, an important future direction is to integrate a more principled and automatic correspondence matching module, for example using local geometric feature descriptors~\cite{Andy20173DMatch,Haowen2018PPFNet,Haowen2018FoldNet} or learning-based matching methods~\cite{wang2019deep,Chen2019NMNet,Christopher2020DGR}.

\subsection{Lack of task-oriented grasp quality}
    The proposed surface-fitting-based grasp planning primarily relies on geometric compatibility and contact stability derived from point cloud observations, and therefore does not explicitly incorporate semantic or contextual information about the downstream task after grasping.
    As a result, task-dependent requirements (e.g., where and from which direction to approach, or which contact region to prioritize) must currently be specified manually, such as by providing an approach-direction vector chosen with heuristics.
    While this is acceptable for controlled settings, it does not scale to larger and more complex scenarios where task-oriented grasp selection is critical.
    From this perspective, a promising direction is to integrate task-driven contact-region or grasp-pose proposals via affordance reasoning~\cite{affordances_IROS_2017,affordance_Humanoids_2017,Denis2022AffCorrs,Toan2023OpenVocabulary,Fan2025Granularity} and vision--language models~\cite{language_CoRL_2023,Toan2024NegativePrompt,Qian2024DexGANGrasp,Chao2025FoundationGrasp}.

\subsection{Limited robustness to incomplete point clouds}
    A further limitation of DISF lies in the assumption that point clouds with sufficiently wide surface coverage are available for grasp planning.
    In our Observed-shape setup, we mitigated this issue by deploying two depth sensors (Orbbec Femto Bolt) at complementary viewpoints and registering/merging the resulting point clouds to obtain a more complete observation of the target geometry.
    However, in many deployments, only a single (often low-cost) depth camera may be available, yielding noisier and more incomplete point-cloud measurements with limited surface visibility.
    This issue becomes even worse in cluttered scenes due to inter-object occlusion, unlike the open-space grasping setting used in our evaluations.
    To extend DISF to such settings, a promising direction is to integrate point-cloud completion techniques~\cite{Xin2020SkipAttention,Xumin2021PoinTr,Xin2021PMPNet,Ruikai2023P2C,Jun2024PointAttN} to reconstruct missing geometry (or grasp-relevant surface regions) from partial observations, enabling robust grasp planning under incomplete point clouds.

\section{Conclusion}
    In this paper, we proposed a novel surface fitting-based grasp planning algorithm that extends conventional geometric compatibility optimization by incorporating CoM alignment to ensure that the gripper and object surfaces are spatially aligned, thereby enhancing contact stability.  
    Inspired by human grasping behavior, our method disentangles the grasp pose optimization process into three sequential steps:  
    (1) rotation optimization to align contact normals,  
    (2) translation refinement for CoM alignment, and  
    (3) gripper aperture adjustment to optimize contact point distribution.  
    We validated DISF through extensive simulations under both Known-shape (clean CAD-derived point clouds) and Observed-shape (YCB point clouds with sensor noise) settings.
    Across these settings, DISF consistently reduced CoM misalignment while maintaining competitive geometric alignment, and this translated into higher grasp success rates in grasp executions.
    We further evaluated cross-platform execution on three robot--gripper platforms, and demonstrated real-world grasp executions on a UR3e, showing that our CoM alignment improved physical feasibility of surface fitting-based grasp planning.

\appendix

\section{Pre-Convex Shape Approximation of Objects for Physical Simulation}
\label{appendix:pre-convex-shape-approx}
When directly loading object model files into the physics simulator for grasping experiments, the object shapes are automatically approximated as convex hulls within the simulator for collision detection purposes.
However, this convex hull approximation is often too coarse, resulting in significant deviations from the original 3D object shape and causing issues with executing grasping experiments accurately.
To address this, a pre-processing step using CoACD~\cite{CoACD} (Collision-Aware Convex Decomposition) was applied to ensure that the original object shape is preserved even after automatic convex approximations are applied within the simulator (in our case, Mujoco 3.2.4). 
The resolution parameter for the CoACD approximation was set to 50.

\section{Pre-Grasp and Grasp Trajectory Planning}
\label{appendix:pre-grasp_and_trajectory}
While the proposed DISF method provides the final grasp pose, it does not plan the complete grasp trajectory, including the intermediate path from the robot's initial pose to the final grasp pose.
Therefore, to execute the planned grasp in practice, an external trajectory planning algorithm is required.
In this experiment, based on the grasp pose obtained from DISF, we first compute a pre-grasp pose and move the robot hand's palm from the initial pose to the pre-grasp pose.

The pre-grasp pose is determined based on the final grasp pose \((\mathbf{R}^*, \mathbf{t}^*)\) provided by DISF. The corresponding palm pose in the world coordinate frame is given as \((\mathbf{R}^*_{\text{palm}}, \mathbf{t}^*_{\text{palm}})\). 
To calculate the pre-grasp pose, we consider a sphere centered at the object grasp position \(\mathbf{t}^*_{\text{palm}}\) with a radius \(\gamma\). 
The pre-grasp pose $\mathbf{p}_{\text{pre}}$ is defined as the surface point on this sphere aligned with the grasp rotation \(\mathbf{R}^*_{\text{palm}}\), computed as:
\begin{equation}
    \label{eq:pre-grasp-position}
    \begin{aligned}
        \mathbf{p}_{\text{pre}} = \mathbf{t}^*_{\text{palm}} - \gamma \cdot \mathbf{R}^*_{\text{palm}} \mathbf{n}_z,
    \end{aligned}
\end{equation}
where \(\gamma\) is the radius of the sphere representing the distance from the grasp position to the pre-grasp position. 
The pre-grasp orientation remains the same as the grasp orientation, represented by the rotation matrix: $\mR_{\textrm{pre}} = \mR^*_{\textrm{palm}}$.     
This formulation ensures the pre-grasp pose is aligned with the intended approach direction, facilitating smooth trajectory planning and grasp execution.

After reaching the pre-grasp pose, the robot continues to move towards the final grasp pose \((\mathbf{R}^*_{\text{palm}}, \mathbf{t}^*_{\text{palm}})\).
Finally, at the final grasp pose, the gripper open/close command is sent to the robot to complete the grasping process using the optimized gripper displacement $\dd^*$.

\section{Fingertip Displacement Refinement for Grasp Execution}
\label{appendix:fingertip_refine}
The planned grasp ensures that the gripper surfaces align with the object surfaces according to the object's geometry. 
However, during actual grasp execution, repulsive forces from the object act against the gripper's grasping force. 
This can result in the gripper failing to firmly hold the object if the planned gripper opening width is used directly. 
To address this issue, we introduce an additional refinement to the gripper's opening width when executing the grasp in the physics simulator. 
Specifically, we add a bias $\hat{\dd}$ to the planned fingertip displacement $\dd$, resulting in a slightly smaller gripper opening width than the planned value. 
This adjustment ensures a more secure grasp during execution by compensating for the effects of repulsive forces.
In both of the simulation and real-world experiment, we applied this refinement process into all three comparative methods.

\begin{table}[t]
  \centering
    \caption{
        Number of object correspondences used for grasp planning across objects and robotic platforms.
        The upper block reports the correspondence counts in simulation (Panda, UR5e, and iiwa) for both the Known-shape and Observed-shape settings.
        The lower block reports the correspondence counts in real-world execution on UR3e, separated into the same objects as the Known-shape setting and additional objects evaluated only in the Observed-shape setting.
    }
  \label{tab:object_correspondences}
  \small
  \setlength{\tabcolsep}{5pt}
  \begin{tabular}{llrrr}
    \toprule
    Setting (\textit{simulation}) & Object & Panda & UR5e & iiwa \\
    \midrule
    \multirow{5}{*}{Known-shape} 
     & \texttt{T-shape\_Block} & 113 & 149 & 531 \\
     & \texttt{Rubber\_Duck} & 32 & 89 & 121 \\
     & \texttt{Hammer} & 34 & 63 & 119 \\
     & \texttt{Wine\_Glass} & 14 & 22 & 36 \\
     & \texttt{Old\_Camera} & 34 & 43 & 203 \\
    \midrule
    \multirow{10}{*}{Observed-shape}
     & \texttt{006\_mustard\_bottle} & 18 & 19 & 141 \\
     & \texttt{011\_banana} & 10 & 30 & 39 \\
     & \texttt{029\_plate} & 16 & 11 & 121 \\
     & \texttt{033\_spatula} & 5 & 14 & 45 \\
     & \texttt{035\_power\_drill} & 14 & 29 & 126 \\
     & \texttt{037\_scissors} & 19 & 31 & 76 \\
     & \texttt{042\_adjustable\_wrench} & 15 & 22 & 100 \\
     & \texttt{052\_extra\_large\_clamp} & 20 & 28 & 68 \\
     & \texttt{058\_golf\_ball} & 23 & 39 & 60 \\
     & \texttt{065\text{-}j\_cups} & 16 & 17 & 58 \\
    \toprule
    Setting (\textit{real world}) & Object &  & UR3e & \\
    \midrule
    \multirow{5}{*}{\makecell[l]{Known-shape\\(same objects)}}
     & \texttt{T-shape\_Block} &  & 73 &  \\
     & \texttt{Rubber\_Duck}   &  & 67 &  \\
     & \texttt{Hammer}         &  & 19 &  \\
     & \texttt{Wine\_Glass}    &  & 10 &  \\
     & \texttt{Old\_Camera}    &  & 74 &  \\
    \midrule
    \multirow{4}{*}{\makecell[l]{Observed-shape\\(additional objects)}}
     & \texttt{Tripod}          &  & 16 &  \\
     & \texttt{USB}             &  & 31 &  \\
     & \texttt{Controller}      &  & 36 &  \\
     & \texttt{Tape}            &  & 52 &  \\
    \bottomrule
  \end{tabular}
\end{table}

\section{Correspondence Number of Objects}
\label{appendix:correspondence_number}

In this appendix, we report the number of object correspondences used VISF and DISF during optimization.
Table~\ref{tab:object_correspondences} summarizes the correspondence number for each (robot, object) combination.
This number depends on both the object geometry and the robot/gripper configuration (e.g., contact area and surface coverage).

\clearpage

\bibliographystyle{elsarticle-num} 
\bibliography{bibtex}


\end{document}